\documentclass[10pt,twocolumn,letterpaper]{article}

\usepackage{cvpr}
\usepackage{times}
\usepackage{epsfig}
\usepackage{graphicx}
\usepackage{amsmath}
\usepackage{amssymb}
\usepackage{multirow}
\usepackage{booktabs}
\usepackage{array}
\usepackage{subfigure}
\usepackage{tabularx}
\usepackage{bbm}
\usepackage{pifont}
\usepackage{xcolor}
\usepackage{titling}

\usepackage[pagebackref=true,breaklinks=true,letterpaper=true,colorlinks,bookmarks=false]{hyperref}

\cvprfinalcopy % *** Uncomment this line for the final submission

\def\cvprPaperID{} % *** Enter the CVPR Paper ID here

\ifcvprfinal\fi
\begin{document}

%%%%%%%%% TITLE
\title{\textbf{DIRV: Dense Interaction Region Voting for End-to-End\\ Human-Object Interaction Detection }}

\author{Hao-Shu Fang$^{1*}$, Yichen Xie$^{1*}$, Dian Shao$^2$, Cewu Lu$^{1\dagger}$\\
$^1$Shanghai Jiao Tong University \quad
$^2$The Chinese University of Hong Kong\\
{\tt\small fhaoshu@gmail.com, xieyichen@sjtu.edu.cn,  sd017@ie.cuhk.edu.hk, lucewu@sjtu.edu.cn}
}

\date{}
\maketitle
\let\thefootnote\relax\footnotetext{* Equal contribution. Names in alphabetical order.}
\let\thefootnote\relax\footnotetext{$\dagger$ Cewu Lu is the corresponding author, member of Qing Yuan Research Institute and MoE Key Lab of Artificial Intelligence, AI Institute, Shanghai Jiao Tong University, China and Shanghai Qi Zhi institute}

\begin{abstract}
Recent years, human-object interaction (HOI) detection has achieved impressive advances. However, conventional two-stage methods are usually slow in inference. On the other hand, existing one-stage methods mainly focus on the \textit{union regions} of interactions, which introduce unnecessary visual information as disturbances to HOI detection. To tackle the problems above, we propose a novel one-stage HOI detection approach \textbf{DIRV} in this paper, based on a new concept called \textit{interaction region} for the HOI problem. Unlike previous methods, our approach concentrates on the densely sampled interaction regions across different scales for each human-object pair, so as to capture the subtle visual features that is most essential to the interaction. Moreover, in order to compensate for the detection flaws of a single interaction region, we introduce a novel \textit{voting strategy} that makes full use of those overlapped interaction regions in place of conventional Non-Maximal Suppression (NMS). Extensive experiments on two popular benchmarks: V-COCO and HICO-DET show that our approach outperforms existing state-of-the-arts by a large margin with the \textit{highest} inference speed and \textit{lightest} network architecture. Our code
is publicly available at \url{www.github.com/MVIG-SJTU/DIRV}.
\end{abstract}

\section{Introduction}
Human-object interaction (HOI) detection aims to recognize and localize the interactions between human-object pairs (\textit{e.g.} sitting on a chair, riding a horse, eating an apple, \textit{etc.}). As a fundamental task of image semantic understanding, it plays a vital role in many other computer vision fields such as image captioning \cite{2019Aligning, DBLP:journals/corr/abs-1808-05864}, visual question answering \cite{2019Relation,DBLP:journals/corr/abs-1806-07243} and
action understanding \cite{CVPR2019_ARG,Shao_2020_CVPR}.

For HOI detection, almost all previous methods emphasized the importance of the \textit{union regions} of an interaction, which covers the whole human, object and intermediate context. For instance, existing two-stage algorithms commonly crop the union region of a human-object pair and then embed its visual features \cite{gupta2018nofrills, gao2018ican, li2019transferable}, while recent one-stage methods aim to regress this union region with keypoints \cite{liao2019ppdm,Wang2020IPNet} or anchor boxes \cite{2020UnionDet} and use it to associate the target human and object.

\begin{figure}[t]
\centering
\subfigure{
    \centering
    \includegraphics[width=0.45\linewidth,height=0.6\linewidth]{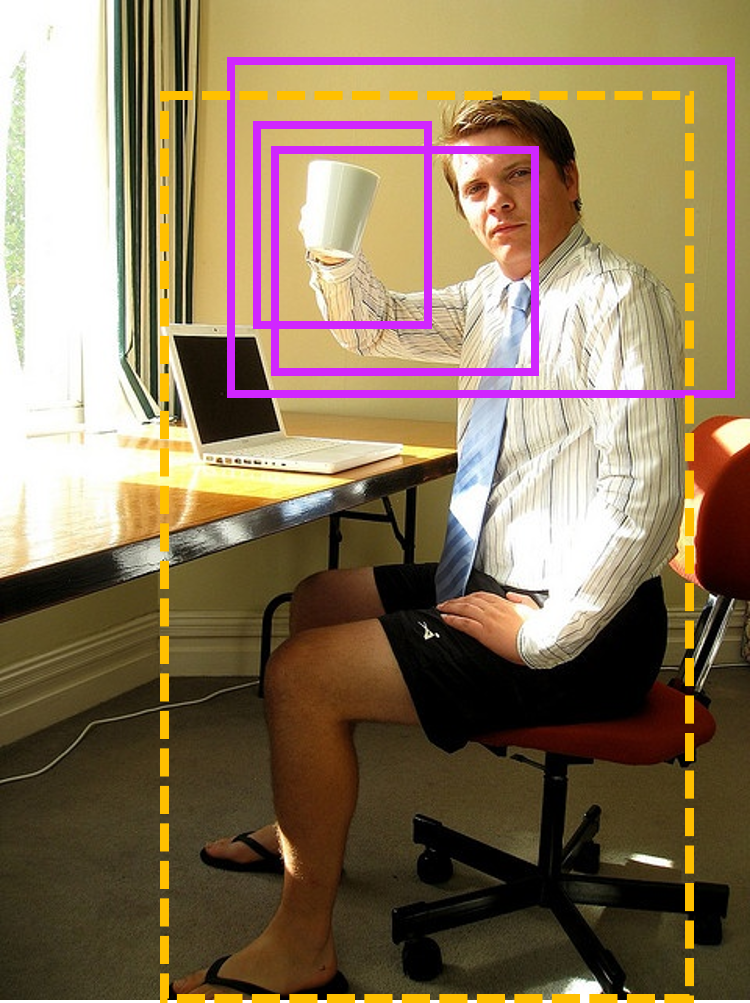}
}
\hspace{0.03\linewidth}
\subfigure{
    \centering
    \includegraphics[width=0.45\linewidth,height=0.6\linewidth]{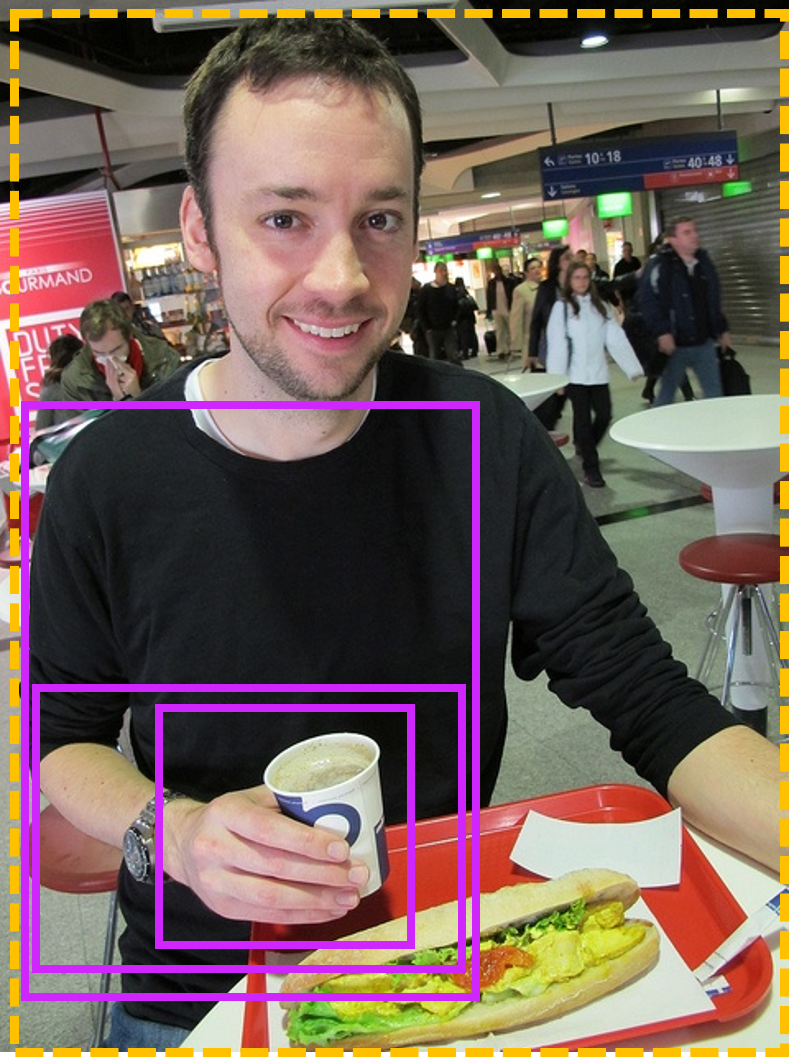}
    \label{fig:interaction region overlap}
}
\caption{\textbf{Union Regions \textit{vs} Interaction Regions:} Conventional approaches usually pays attention to the \textit{union region} (\textit{dashed yellow}), which contains too much redundant information. Instead, we propose a method focusing on \textit{interaction regions} (\textit{solid violet}) with different scales. In above two figures, despite distinct human/object poses, \textit{interaction regions} cover the most critical segments containing the cups, hands or arms, when detecting \textit{holding a cup}.}
\label{fig:interactive_region}
\vspace{0.0in}
\end{figure}

However, we find that such emphasis on union region is \textit{counter-intuitive} for human beings. In practice, it is not necessary to observe the whole union region before making decisions in most situations. For instance, when asked to determine whether a man is holding a cup, we only need to notice his hands but never care about where his feet are. That's to say, humans can easily target the human-object pair of an HOI, without the needs of being told the union regions. Based on these observations, we propose a new recognition unit for HOI detection, called \textit{interaction region}. The interaction region denotes the region that covers the minimal area of human and object crucial for recognizing the interaction.  An example is given in Fig.~\ref{fig:interactive_region}.   In this case, an upper-body region that contains a cup and hand would be more distinguishable than the union region.

To this end, we propose a novel one-stage HOI detector that concentrates on the interaction regions of human-object interactions.
We hypothesize that these regions are highly informative to determine the interaction category and human-object relative spatial configuration. To fully utilize the interaction regions for HOI detection, three main technical challenges identified as follows need to be addressed beforehand.

\textit{Challenge 1: How do we decide the interaction regions?} Although recent work provided part-level action labels~\cite{li2020pastanet}, we tend to seek a more general and simpler HOI detector without the need for extra annotations.  Empirically, we consider that those human parts closer to the object are more likely to take an indispensable effect on the interaction, and so are the object parts. For simplicity, we consider some rectangle regions, which cover both some parts of the human and object, as interaction regions. A natural idea comes by applying the dense anchor boxes in one-stage object detection models to represent these regions. To achieve that, we set three overlapping thresholds between anchor boxes and human bounding boxes, object bounding boxes as well as union regions. We apply a dense interaction region selection manner, where all anchors satisfying these three thresholds are regarded as interaction regions.

\textit{Challenge 2: An anchor box may be regarded as the interaction region for multiple different HOIs.} Unlike object detection, this situation appears frequently in HOI detection. Under this condition, the anchor box needs to predict multiple HOI labels and corresponding object locations, where the number is unfixed. This poses extra challenges for network design and final result association. Therefore, we match each anchor box with only one unique interaction. In addition, there inevitably exists some missed positive interactions within the popular datasets. We develop a novel \textit{ignorance loss} based on classical focal loss \cite{DBLP:journals/corr/abs-1708-02002} to address these problems.

\textit{Challenge 3: Single interaction region may lead to ambiguity or misrepresentation.} HOI recognition relies on very subtle visual cues in interaction regions. Some visual features are even ambiguous, leading to the fragile result from a single anchor. For this reason, we propose a novel \textit{voting strategy}. Each anchor only contributes a little to the final location and classification prediction. For each interaction type, a \textit{probability distribution} is established for the relative location between each human-object pair by fusing the prediction results of different anchors. This \textit{dense anchor voting strategy} can remarkably elevate the fault-tolerance of each anchor and achieve a robust final prediction.

Extensive experiments show that our one-stage approach, \textbf{DIRV} (\textbf{D}ense \textbf{I}nteraction \textbf{R}egion \textbf{V}oting), outperforms existing state-of-the-art models on two popular benchmarks, achieving both \textit{higher accuracy} and \textit{faster speed}.

\iffalse
As shown in Fig.~\ref{fig:mAP_time_params}, it achieves both \textit{higher accuracy} and \textit{faster speed} and requires no extra annotations.

Our contributions are summarized as below:

\begin{itemize}
    \item We propose a new concept called \textit{interaction region} for HOI detection problem. Our dense sampling strategy naturally captures the crucial regions of HOI and encodes multi-scale features.
    \item We propose the \textit{dense anchor voting strategy} to improve the fault-tolerance of single anchor's result. This can be easily extended to other visual relationship detection tasks.
    \item We develop a meta network architecture, \textit{DIRV}, which is significantly more simple and intuitive than previous work. It notably surpasses current state-of-the-art methods on V-COCO and HICO-DET with a very high time and space efficiency.
\end{itemize}
\fi

\section{Related Work}
%\subsection{Human-Object Interaction Detection}
Human-object interaction (HOI) detection is formally defined as retrieving \textit{$\langle$human, verb, object$\rangle$} triplets from images. Previous methods mainly employed a two-stage strategy. In the first stage, a pre-trained object detector \cite{DBLP:journals/corr/LinDGHHB16,DBLP:journals/corr/RenHG015} localized both humans and objects within the image. In the second stage, a classification network recognized the interaction categories for each human-object pair. Most work focused on the improvement of the second stage. Some early work \cite{gupta2015visual} simply extracted features from each human or object instance. This method suffered from lack of contextual information. Afterwards, more information was taken into account rather than instance appearance, including spatial location \cite{chao2018learning,gao2018ican, qi2018learning}, human pose \cite{fang2018pairwise, li2019transferable,wan2019pose}, word embedding \cite{DBLP:journals/corr/abs-1904-03181,lu2016visual}, segmentations \cite{zhou2020cascaded, fang2018weakly} and human part label \cite{li2020pastanet}. Yet, these two-stage methods typically need to detect all human-object pairs, making their inference time grow quadratically with instance number. Furthermore, these approaches usually adopted a heavy network for classification, which led to considerable computation overhead.

To tackle these drawbacks, some recent work developed one-stage HOI detectors. Liao \textit{et.al.} \cite{liao2019ppdm} and Wang \textit{et.al.} \cite{Wang2020IPNet} posed HOI detection as a keypoint detection and grouping problem. Despite their impressive efficiency and accuracy, the interaction keypoints had no apparent characteristics in visual patterns so the networks were not easy to train. Kim \textit{et.al.} \cite{2020UnionDet} designed an anchor-based one-stage algorithm to regress the union region of human and object. However, as aforementioned, union region prediction is not straight-forward and single anchor's prediction is fragile.

Unlike all the above methods, our method makes full use of visual patterns within interaction regions across different scales, allowing a promising accuracy without the help of any other proposals or annotations. The one-stage strategy and concise network architecture also bring greatly improved running time and space efficiency.

\if0
\subsection{One-Stage Object Detection}
One-stage object detection algorithms \cite{DBLP:journals/corr/LiuAESR15,DBLP:journals/corr/abs-1904-01355,2019EfficientDet} were proposed in the pursuit of faster inference speed. These detectors conducted classification and regression on all anchor boxes in each position. Our one-stage HOI detection mechanism is general and can be built upon any popular backbone.
\fi

\section{Methods}
In this section, we introduce our proposed \textbf{DIRV} (Dense Interaction Regions Voting) framework for human-object interaction (HOI) detection. The problem formulation is firstly explained in Sec.~\ref{sec:formulation}. Then, we present the network architecture of our detector in Sec.~\ref{sec:detector}. Afterwards, the inference protocol based on \textit{voting strategy} is shown in Sec.~\ref{sec:inference}. Finally, we demonstrate how to train our deep neural network model in Sec.~\ref{sec:training}.

\begin{figure*}
    \centering
    \includegraphics[width=0.9\linewidth]{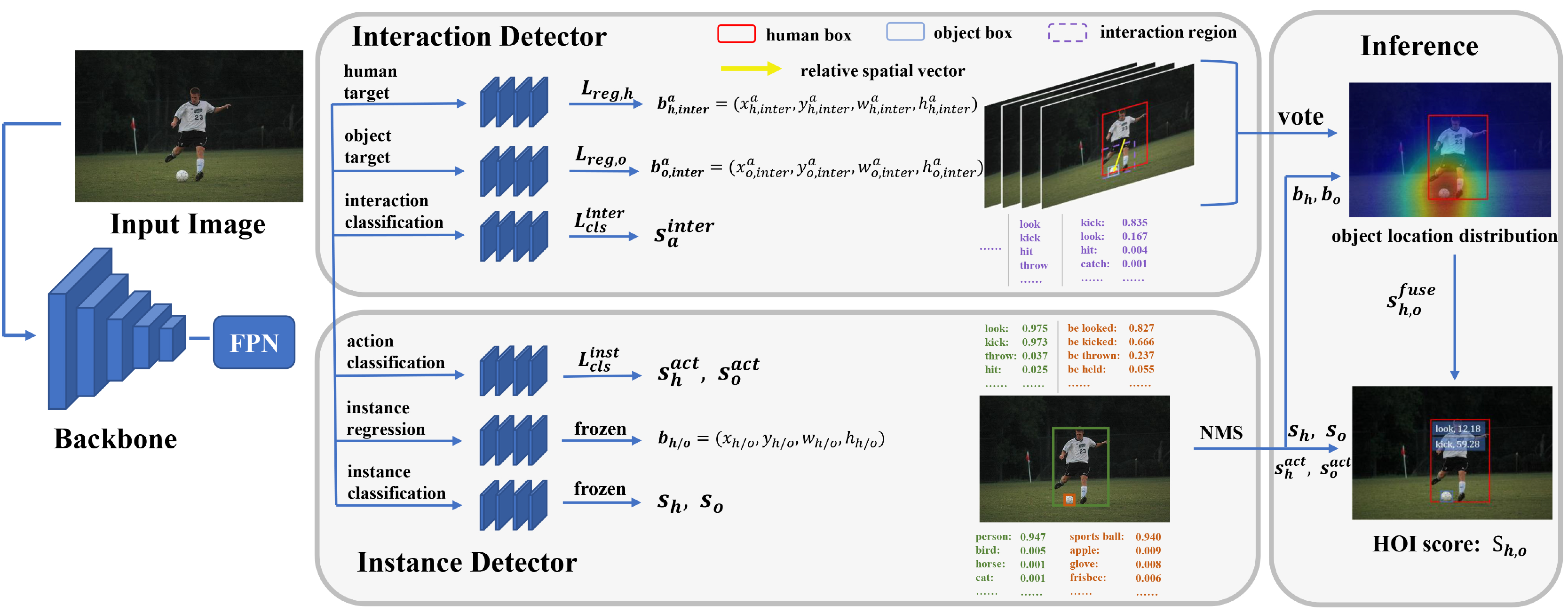}
    \caption{\textbf{Overview of our DIRV Framework:} It is composed of two components: \textit{Interaction Detector} and \textit{Instance Detector}. For each interaction region, a relative spatial vector is obtained by regressing the human and object bounding boxes. During inference, results of interaction regions vote for an \textit{object location distribution}, from which HOI score is derived.}
    \label{fig:overview}
\vspace{0.0in}
\end{figure*}

\subsection{Formulation}
\label{sec:formulation}
Typically, HOI detection aims to fetch a \textit{$\langle b_h,v,b_o\rangle$} triplet for each interaction within a single image $x$, where $b_h,b_o$ denote the bounding box of human $h$ and object $o$ separately, while $v$ denotes the human action. Without considering external input like human poses~\cite{fang2017rmpe}, conventional two-stage HOI detectors formulate the problem as

\begin{equation}
    \begin{aligned}
    &\mathcal{H},\mathcal{O} = d(\mathbf{f_x}),\\
    &v_i = g(b_h, b_o, \mathbf{f_x}), \forall h \in \mathcal{H}, \forall o \in \mathcal{O},
    \end{aligned}
\end{equation}
where $d(\cdot)$ is a vanilla object detector, $g(\cdot)$ is the verb classifier for a human-object pair, $\mathbf{f_x}$ is the appearance feature of the whole image $x$ and $\mathcal{H}$,$\mathcal{O}$ are detected humans and objects. Since the input of $g(\cdot)$ relies on the output of $d(\cdot)$, these two processes cannot run in parallel and $g(\cdot)$ would face the combinatorial explosion problem. On the contrary, we reformulate HOI detection as
\begin{equation}
    \begin{aligned}
    &\mathcal{H},\mathcal{O} = d(\mathbf{f_x}),\\
    &\langle T(b_h), v, T(b_o)\rangle = g(\mathbf{f_x}), h \in \mathcal{H},  o \in \mathcal{O},
    \end{aligned}
\end{equation}
where $T(\cdot)$ is a target indicator that links the verb to a detected human-object pair. By doing so, we can run these two processes simultaneously.

Further, we do not adopt the common practice of Non-Maximum Suppression (NMS) when retrieving the $\langle T(b_h), v, T(b_o)\rangle$. In contrast, we propose a different strategy, \textit{voting}, to handle the prediction of different \textit{interaction regions}. Predictions based on every anchor's visual features are fully utilized instead of being suppressed. The final HOI prediction comes from the combination of each interaction region through voting. To sum up, our algorithm is formulated as Eq.~\ref{eq:voting}:

\begin{equation}
    \begin{aligned}
    &\mathcal{H},\mathcal{O} = d(\mathbf{f_x}),\\
        & \langle T(b_h^i), v^i, T(b_o^i)\rangle = g(\mathbf{f_x^{a_i}}),i\in\{1,2,\dots,N\},\\
        &\langle T(b_h), v, T(b_o)\rangle=vote(\{ \langle T(b_h^i), v^i, T(b_o^i)\rangle\}_{i\in\{1,\dots,N\}}),
    \end{aligned}
    \label{eq:voting}
\end{equation}
where $\langle T(b_h^i), v^i, T(b_o^i)\rangle$ is the prediction based on anchor $a_i$. $N$ is the number of interaction regions for this interaction. We show how we obtain  $\mathcal{H},\mathcal{O}$ and $\langle T(b_h^i), v^i, T(b_o^i)\rangle$ for each anchor in Sec.~\ref{sec:detector}. $vote(\cdot)$ is the voting strategy, which is elaborated in Sec.~\ref{sec:inference}.

\subsection{Dense Interaction Region Detector}
\label{sec:detector}
Our network structure is illustrated in Fig.~\ref{fig:overview}.  The model is composed of two components: an instance detector and an interaction detector. Each of them contains three parallel sub-branches, which share the feature map of the Feature Pyramid Network. We first explain the instance detector for $\mathcal{H},\mathcal{O}$ and then the interaction detector for $\langle T(b_h^i), v^i, T(b_o^i)\rangle$.

\subsubsection{Instance Detector}
The instance detector mainly helps instance localization and supports the detection of none object actions, \textit{e.g.} \textit{walking}. It contains three sub-branches: \textit{instance classification branch}, \textit{instance regression branch} and \textit{instance action classification branch}.

The instance regression and classification branches follow the standard setting in most object detection networks, which regress instance bounding boxes based on anchors as well as classify these instances. Interactions are not considered in these two branches.

Beyond these two branches, an instance action classification branch plays an auxiliary role in interaction classification. It predicts the action scores of humans and objects, helping the association of human-verb-object pair. The actions of humans and objects are treated separately, \textit{e.g.}, \textit{hold} and \textit{be held} are classified as two different actions. If there are $C_h$ human actions and $C_o$ object actions, the classification gives two scores $s_h^{act}\in\mathbb{R}^{C_h}$ and $s_o^{act}\in\mathbb{R}^{C_o}$. The anchor settings follow standard object detection and only those positive anchors involved in at least one interaction are taken into account when calculating loss.

\subsubsection{Interaction Detector}

The interaction detector serves as the key of our proposed architecture, \textbf{DIRV}. It directly predicts the interaction $v^i$ and the target $\langle T(b_h^i), T(b_o^i)\rangle$ that indicates the corresponding human-object pair from the subtle visual features in \textit{interaction regions}. We first clarify our \textit{methodology}, followed by two key learning techniques: \textit{interaction region decision} and \textit{ignorance loss}.
\vspace{0in}
\paragraph{Methodology:} To retrieve the $\langle T(b_h^i), v^i, T(b_o^i)\rangle$ triplet, we design three parallel sub-branches: \textit{interaction classification branch}, \textit{human target branch} and \textit{object target branch} for predicting $v^i, T(b_h^i),$ and $T(b_o^i)$ separately.

    The interaction classification branch classifies the interaction type $v^i$ within the interaction region (\textit{i.e.} the anchor). It obtains an interaction score prediction $s_{a_i}^{inter}\in\mathbb{R}^C$ for each interaction region $a_i$, where $C$ is the number of interaction categories.

    For human and object targets $T(b_h^i)$ and $T(b_o^i)$, it is difficult to directly link the verb to the detected human and object given by the \textit{instance detector} since the detection branch run in parallel. Thus, we propose an intuitive yet effective solution. The human target branch regresses the human bounding box $b_{h,inter}^{a_i}=(x_{h,inter}^{a_i},y_{h,inter}^{a_i},w_{h,inter}^{a_i},h_{h,inter}^{a_i})$ from the anchor $b_a^{a_i}=(x_a^{a_i},y_a^{a_i},w_a^{a_i},h_a^{a_i})$, where $(x_{h,inter}^{a_i},y_{h,inter}^{a_i})$ is its bounding box center. Similarly, the object target branch regresses the object bounding box $b_{o,inter}^{a_i}=(x_{o,inter}^{a_i},y_{o,inter}^{a_i},w_{o,inter}^{a_i},h_{o,inter}^{a_i})$. These predicted human and object bounding boxes serve as the target indicators $T(b_h^i)$ and $T(b_o^i)$. We can easily link the verb $v^i$ to the detected human and object box $b_h^i, b_o^i$ during inference via simple post processing (\textit{e.g.,} IoU matching), which is introduced in Sec.\ref{sec:inference}.

\paragraph{Interaction Region Decision:} As explained before, the interaction regions should cover both parts of interacting human and object. With different scales, these regions may provide important visual features of different levels. Interestingly, we find that such a setting naturally matches the characteristic of anchor boxes $\mathcal{A}$. An anchor box $a_j\in\mathcal{A}$ serves as an interaction region of interaction $I_i$ so long as it satisfies the following overlapping requirement:

    \begin{equation}
        \begin{aligned}
            &O_i^j=\mathbbm{1}\left(IoU(a_j,\hat{b}_u^i)>t_u\right)\cdot\\
            &\mathbbm{1}\left(\frac{a_j\cap \hat{b}_h^i}{\hat{b}_h^i}>t_h\right)\cdot\mathbbm{1}\left(\frac{a_j\cap \hat{b}_o^i}{\hat{b}_o^i}>t_o\right)
        \end{aligned}
        \label{eq:overlapping}
    \end{equation}
    where $\hat{b}_h^i,\hat{b}_o^i$ are the ground-truth human/object bounding box of a possible interaction pair $I_i$. $\hat{b}_u^i$ is the union region box of interaction $I_i$, which is the smallest box that completely covers both $\hat{b}_h^i,\hat{b}_o^i$. $t_u,t_h,t_o$ are three thresholds. We set them as $t_u=t_h=t_o=0.25$, which is analyzed in ablation study.

    With the requirement above, single anchor box may serve as the interaction region of multiple interactions, which impedes the human/object regression. Thus, we define a \textit{overlapping level} metric to ensure that an anchor box corresponds to at most a unique interaction, \textit{i.e.,}
    \begin{equation}
        \begin{aligned}
            \hat{O}_i^j=IoU(a_j,\hat{b}_u^i)+\sqrt{\frac{a_j\cap \hat{b}_h^i}{\hat{b}_h^i}\cdot\frac{a_j\cap \hat{b}_o^i}{\hat{b}_o^i}}.
        \end{aligned}
    \end{equation}
    If multiple interactions are matched with the same anchor box, it will associate with interaction $I_k$ where $\hat{O}_k^j=\max\limits_i\left\{\hat{O}_i^j|O_i^j=1\right\}$ so each anchor has at most one ground-truth in regression.

\paragraph{Ignorance Loss:} For human/object target branch, we just follow many anchor-based object detection methods to apply the standard smooth $L_1$ loss between predicted $b_{h,inter}^{a_i}/b_{o,inter}^{a_i}$ and ground-truth $\hat{b}_h^i/\hat{b}_o^i$ on their loss functions $\mathcal{L}_{reg,h}/\mathcal{L}_{reg,o}$ for interaction region $a_i$.

    Yet, standard focal loss is not applicable for interaction classification branch because of the following two reasons: Firstly, the receptive field of an anchor may contain multiple different interactions. Secondly, HOI detection datasets have much more missed positive samples than object detection datasets. These cause serious confusion during training.

    We propose a novel \textit{ignorance loss} based on vanilla focal loss \cite{DBLP:journals/corr/abs-1708-02002} to address both difficulties above. We eliminate the influence of missed unlabelled interactions by removing the background loss \textit{i.e.} anchors associated with none interactions \textit{don't} take effect in learning.

    Further, as a solution to the multiple interactions problem, we modify the ground-truth targets of foreground anchors as below. For anchor $a_j$, if there exist multiple interactions $\{I_i\}$ within current anchor where $O_i^j=1$, we set the target label as

    \begin{equation}
        t_j^c=
        \begin{cases}
            1 & I_k^c=1, \hat{O}_k^j=\max\limits_i\{\hat{O}_i^j|O_i^j=1\}\\
            0 & I_i^c=0, \forall i, O_i^j=1\\
            ignored & others%I_k^c=0, \exists l\neq k,O_l^j=1: I_l^c=1
        \end{cases}
    \end{equation}
    where $t_j^c$ is the target label of interaction category $c$ for anchor $a_j$. $I_i^c=1$ denotes interaction $I_i$ is positive for category $c$, else $I_i^c=0$. The above equation means that we ignore the classification loss for those interaction categories exist but not dominant in an anchor.

\subsection{Voting Based Model Inference}
\label{sec:inference}
Our model makes inference by combining the prediction results of different interaction regions. Each interaction region contributes to the final interaction recognition with the \textit{weighted localization score} as weight. The inference process is divided into three steps as follows.

\subsubsection{Parallel Inference}
All six sub-branches work in parallel during inference, which dramatically reduces the inference time. From \textit{instance detector}, a set of human $\mathcal{H}$ and object $\mathcal{O}\ (\mathcal{H}\subset\mathcal{O})$ candidates are generated after NMS. For each human instance, we get its bounding box $b_{h}\in\mathbb{R}^4$, instance classification score $s_{h}\in\mathbb{R}$ and instance action classification score $s_{h}^{act}\in\mathbb{R}^{C_h}$. $s_{h}\in\mathbb{R}$ is a scalar since an instance can only be classified as a unique object category with highest score (here is \textit{human}) while $s_{h}^{act}\in\mathbb{R}^{C_h}$ is a $C_h$-d vector. Similarly, we obtain bounding box $b_{o}\in\mathbb{R}^4$, instance classification score $s_{o}\in\mathbb{R}$ and instance action classification score $s_{o}^{act}\in\mathbb{R}^{C_o}$ for each object.

In \textit{interaction detector}, it fetches a triplet of $(b_{h,inter}^{a_i},s_{a_i}^{inter},b_{o,inter}^{a_i})$ from each interaction region $a_i$, where $b_{h,inter}^{a_i},b_{o,inter}^{a_i}\in\mathbb{R}^4$ are the human/object target bounding boxes and $s_{a_i}^{inter}\in\mathbb{R}^C$ is the interaction classification score for each interaction region. Here, we should have $C=C_h=C_o$ after eliminating interactions with none objects.

\subsubsection{Object Location Estimation}
We retrieve the $\langle b_h, v, b_o\rangle$ triplet in a human-centric manner. For each interaction region $a_j$, we first try to match it with a human instance $h^{a_j}\in\mathcal{H}$ based on the overlapping metric, that is

\begin{equation}
    \begin{aligned}
    &IoU(a_j,b_h^{a_j})  =\max\limits_h IoU(a_j,b_h),\\
    &h\in\mathcal{H}, \frac{a_j\cap b_h}{b_h}>t_h
    \end{aligned}
    \label{eq:anchor human}
\end{equation}
where $b_h$ is the human bounding box and $t_h$ is the threshold same as that in Eq.~\ref{eq:overlapping}. If no human instance meets the requirement, this interaction region is abandoned.

After matching the interaction region to a detected human instance, we then search its corresponding object instance. A natural thinking is to match the object like Eq.\ref{eq:anchor human}. However, we found that the location of object is usually not accurate enough. To improve the robustness, we build a probability distribution for the object location based on the prediction result. Referring to \cite{Gkioxari_2018_CVPR}, we model it with a 2-d Gaussian distribution:

\begin{equation}
    p_{a_j}(x_o,y_o)=e^{-\frac{\left|\left|v_{o|h}^{a_j}-\mu^{a_j}_{o|h}\right|\right|^2}{2\cdot \sigma^2}}
    \label{eq:gaussian}
\end{equation}
where $v_{o|h}^{a_j}$ and $\mu_{o|h}^{a_j}$ are the relative object locations scaled by anchor width and height:
\begin{equation}
\begin{aligned}
    &v_{o|h}^{a_j}=\left(\frac{x_o-x_h^{a_j}}{w_a^{a_j}},\frac{y_o-y_h^{a_j}}{h_a^{a_j}}\right),\\
    &\mu_{o|h}^{a_j}=\left(\frac{x_{o,inter}^{a_j}-x_{h,inter}^{a_j}}{w_a^{a_j}},\frac{y_{o,inter}^{a_j}-y_{h,inter}^{a_j}}{h_a^{a_j}}\right),
\end{aligned}
\end{equation}
and the standard deviation $\sigma$ is a hyper-parameter, which is set as $0.9$ in our experiments. As analyzed in the   supplementary material, our method is insensitive to $\sigma$.

After obtaining the object location distribution, we weight it by \textit{interaction classification score} $s_{a_j}^{inter}$ as below.

% For each interaction region, it contributes to the object location prediction by the weight of \textit{interaction classification score} $s_{a_j}^{act}$. We define its \textit{weighted localization score} as the weighted distribution probability:
\begin{equation}
    s_{a_j}^{loc}(x_o,y_o) = s_{a_j}^{inter} \cdot p_{a_j}(x_o,y_o)
\end{equation}
where $(x_o,y_o)$ is the center of object bounding box. Until now, we obtain the \textit{weighted localization scores} $s_{a_j}^{loc}(x,y)\in\mathbb{R}^C$ for all $C$ interaction categories.

\begin{figure}[t]
    \centering
    \subfigure[eat]{
        \includegraphics[width=0.45\linewidth, height=0.4\linewidth]{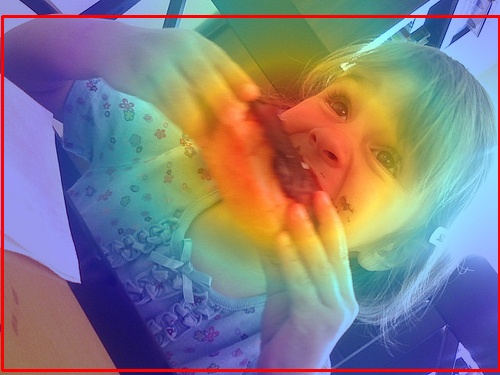}
    }
    \subfigure[talk on phone]{
        \includegraphics[width=0.45\linewidth, height=0.4\linewidth]{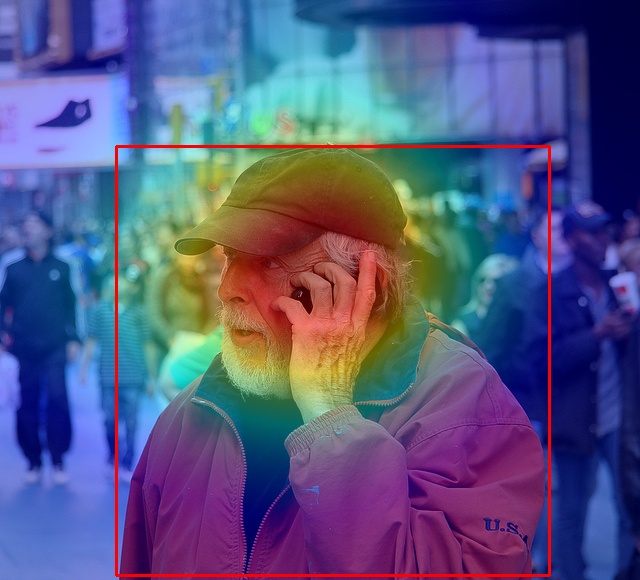}
    }

    \subfigure[surf]{
        \includegraphics[width=0.45\linewidth, height=0.4\linewidth]{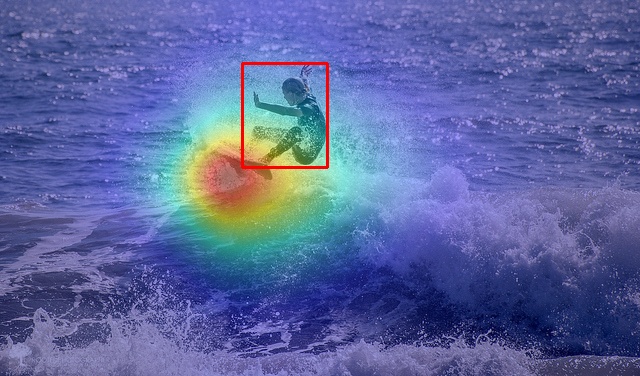}
    }
    \subfigure[throw]{
        \includegraphics[width=0.45\linewidth, height=0.4\linewidth]{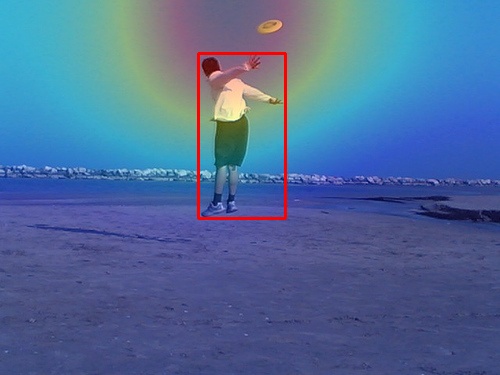}
    }
    \caption{\textbf{Object Location Distribution:} we visualize the target object location distribution for some human instances of several categories. Our voting strategy accurately localizes the objects in these interactions.}
    \label{fig:obj_loc}
\vspace{0in}
\end{figure}

\subsubsection{Voting Based Region Fusion}
By fusing \textit{weighted localization scores} of interaction regions associated with same human instance $b_h$, a \textit{human-centric object location distribution} $s_h^{fuse}$ is computed with our voting strategy:
\begin{equation}
    s_h^{fuse}(x,y)= \sum_{a_j\in\mathcal{A}_h}s_{a_j}^{loc}(x,y),
    \label{eq:location score}
\end{equation}
where $\mathcal{A}_h=\{a_j\}_{h^{a_j}=h}$ is set of interaction regions associated with human instance $h$. We visualize some examples of the fused distribution in Fig.~\ref{fig:obj_loc}.

Finally, we are now able to score a human-object pair using this distribution. For each interaction region, we first associate it with a detected object instance $o^{a_j}$, like Eq.~\ref{eq:anchor human}.

\begin{equation}
    \begin{aligned}
        &p_{a_j}(x_o^{a_j},y_o^{a_j})=\max\limits_o p_{a_j}(x_o,y_o),\\
        &o^{a_j}\in\mathcal{O},\frac{a_j\cap b_o}{b_o}>t_o.
    \end{aligned}
\end{equation}
Then, Eq.~\ref{eq:location score} is rewritten for each specific human-object pair.
\begin{equation}
    s_{h,o}^{fuse}= \sum_{a_j\in\mathcal{A}_{h,o}}s_{a_j}^{loc}(x_o,y_o)
\end{equation}
where $\mathcal{A}_{h,o}$ denotes all the interaction regions $\{a_j\}$ associated human-object pair $(b_h,b_o)$ where $(b_h,b_o)=(b_h^{a_j},b_o^{a_j})$.
Thus, the final HOI score for a human-object pair $(b_h,b_o)$ can be derived as

\begin{equation}
    S_{h,o} = s_h\cdot s_o \cdot (s_h^{act} + s_o^{act}) \cdot s_{h,o}^{fuse}
    \label{eq:score}
\end{equation}
where $s_h,s_o,s_h^{act},s_o^{act}$ have been explained in section \textit{Parallel Inference}. When no object is involved, we simply define $S_h=s_h\cdot s_h^{act}$. The HOI scores are not normalized because we only care about their relative value for the same interaction category.

The time complexity of voting is $O(\left|\mathcal{A}_{pos}\right|)$, where $\mathcal{A}_{pos}=\mathop{\cup}\limits_{h,o}\mathcal{A}_{h,o}$ is the set consisting of all interaction regions associated with any interactive human-object pairs. The size is not very large and it is easy to compute in parallel, so only a little CPU overhead is introduced.

\subsection{Model Training}
\label{sec:training}
During training, the backbone, feature pyramid network and instance classification/regression branches are frozen with COCO pre-trained weight \cite{2019EfficientDet}. The final loss is the sum of loss functions for other four sub-branches in Fig.~\ref{fig:overview}.

\begin{equation}
    \mathcal{L}=\mathcal{L}_{reg,h}+\mathcal{L}_{reg,o}+\mathcal{L}_{cls}^{inter}+\mathcal{L}_{cls}^{inst}
\end{equation}
In \textit{interaction detector}, $\mathcal{L}_{reg,h},\mathcal{L}_{reg,o}$ are the smooth $L_1$ losses for \textit{human and object target branches} separately. $\mathcal{L}_{cls}^{inter}$ is our \textit{ignorance loss} for \textit{interaction classification branch}. We follow focal loss \cite{DBLP:journals/corr/abs-1708-02002} to set $\alpha=0.25,\gamma=2.0$. In \textit{instance detector}, $\mathcal{L}_{cls}^{inst}$ is standard binary cross-entropy loss for \textit{instance action classification branch}.

\begin{table*}[ht!]
    \centering
    \caption{\textbf{Results on V-COCO:}\textit{Proposal} shows whether it needs object detection beforehand. For \textit{Additional}, \textit{P,B,L} denotes human pose, human body part states and language priors respectively, which are utilized in prior methods.}
    \label{tab:vcoco}
    \begin{tabular}{lccccc}
        \toprule
        \textbf{Method} & \textbf{Proposal} & \textbf{Additional} & $\textbf{mAP}_{role}$ & \textbf{Inference Time} & \textbf{Params}\\
        \midrule
        $\textcolor{gray}{RP_DC_D}$ \textcolor{gray}{\cite{li2019transferable}} & \textcolor{gray}{\ding{51}} & \textcolor{gray}{P} & \textcolor{gray}{47.8} & \textcolor{gray}{513 ms} & \textcolor{gray}{64 M} \\
        \textcolor{gray}{PMFNet \cite{wan2019pose}} & \textcolor{gray}{\ding{51}} & \textcolor{gray}{P} & \textcolor{gray}{52.0} & \textcolor{gray}{253 ms} & \textcolor{gray}{179 M}\\
        \textcolor{gray}{ConsNet \cite{2020arXiv200806254L}} & \textcolor{gray}{\ding{51}} & \textcolor{gray}{P+L} & \textcolor{gray}{53.2} &\textcolor{gray}{-} & \textcolor{gray}{-}\\
        \textcolor{gray}{MLCNet \cite{10.1145/3372278.3390671}} & \textcolor{gray}{\ding{51}} & \textcolor{gray}{P+B+L} & \textcolor{gray}{55.2} &\textcolor{gray}{-} &\textcolor{gray}{-} \\
        \midrule
        InteractNet \cite{Gkioxari_2018_CVPR} & \ding{51} &\ding{55} & 40.0 & 145 ms & 71 M \\
        iCAN \cite{gao2018ican} & \ding{51}&\ding{55} & 44.7 & 204 ms & 89 M\\
        Zhou \textit{et.al.} \cite{zhou2020cascaded} & \ding{51} & \ding{55} &48.9 & - & 620 M  \\
        VSGNet \cite{2020VSGNet} & \ding{51} & \ding{55} & 51.8 & 312 ms & 59 M \\
        UnionDet \cite{2020UnionDet} & \ding{55} & \ding{55} & 47.5 & 78 ms & 44 M\\
        IP-Net \cite{Wang2020IPNet} & \ding{55} & \ding{55} & 51.0 &-& - \\
        \midrule
        \textbf{DIRV (ours)} & \ding{55} & \ding{55} & \textbf{56.1} & \textbf{68 ms} & \textbf{12 M}\\
        \bottomrule
    \end{tabular}

\end{table*}

\begin{table*}[ht!]
    \centering
    \caption{\textbf{Results on HICO-DET:} \textit{Proposal} shows whether it needs object detection beforehand. For \textit{Additional}, \textit{P,B,L} denotes human pose, human body part states and language priors respectively, which are utilized in prior methods.}
    \label{tab:hico}
    \resizebox{\textwidth}{!}{
    \begin{tabular}{lcccccccccc}
        \toprule
        \multirow{2}{*}{\textbf{Method}} & \multirow{2}{*}{\textbf{Proposal}} & \multirow{2}{*}{\textbf{Additional}} &  \multicolumn{3}{c}{\textbf{Default}} & \multicolumn{3}{c}{\textbf{Known Object}} & \multirow{2}{*}{\textbf{Inference Time}} & \multirow{2}{*}{\textbf{Params}}\\
        & & & \textbf{Full} & \textbf{Rare} & \textbf{Non-Rare} & \textbf{Full} & \textbf{Rare} & \textbf{Non-Rare}\\
        \midrule
        $\textcolor{gray}{RP_DC_D}$ \textcolor{gray}{\cite{li2019transferable}} & \textcolor{gray}{\ding{51}} & \textcolor{gray}{P} &\textcolor{gray}{17.03} & \textcolor{gray}{13.42} & \textcolor{gray}{18.11} & \textcolor{gray}{19.17} & \textcolor{gray}{15.51} & \textcolor{gray}{20.26} & \textcolor{gray}{513 ms}& \textcolor{gray}{64 M}\\
        \textcolor{gray}{PMFNet \cite{wan2019pose}} & \textcolor{gray}{\ding{51}} & \textcolor{gray}{P} & \textcolor{gray}{17.46} & \textcolor{gray}{15.65} & \textcolor{gray}{18.00} & \textcolor{gray}{20.34} &\textcolor{gray}{17.47} & \textcolor{gray}{21.20} & \textcolor{gray}{253 ms}& \textcolor{gray}{179 M}\\
        \textcolor{gray}{MLCNet \cite{10.1145/3372278.3390671}} & \textcolor{gray}{\ding{51}} & \textcolor{gray}{P+B+L} &\textcolor{gray}{17.95} &\textcolor{gray}{16.62} &\textcolor{gray}{18.35} &\textcolor{gray}{22.28} &\textcolor{gray}{20.73} &\textcolor{gray}{22.74} &\textcolor{gray}{-} &\textcolor{gray}{-}\\
        % \textcolor{gray}{Functional \cite{DBLP:journals/corr/abs-1904-03181}} & \textcolor{gray}{\ding{51}} & \textcolor{gray}{L} & \textcolor{gray}{21.96} & \textcolor{gray}{16.43} & \textcolor{gray}{23.62} & \textcolor{gray}{-} & \textcolor{gray}{-} & \textcolor{gray}{-}\\
        \textcolor{gray}{ConsNet \cite{2020arXiv200806254L}} & \textcolor{gray}{\ding{51}} & \textcolor{gray}{P+L} &\textcolor{gray}{22.15} &\textcolor{gray}{17.12} & \textcolor{gray}{23.65} &\textcolor{gray}{-} &\textcolor{gray}{-} &\textcolor{gray}{-} &\textcolor{gray}{-} &\textcolor{gray}{-}\\
        \midrule
        InteractNet \cite{Gkioxari_2018_CVPR} & \ding{51} &\ding{55} & 9.94& 7.16&10.77 &- &- &- & 145 ms & 72 M \\
        iCAN \cite{gao2018ican} & \ding{51}&\ding{55} &14.84 &10.45 &16.15 &16.26 &11.33 & 17.73 & 204 ms & 89 M\\
        UnionDet \cite{2020UnionDet} & \ding{55} & \ding{55} & 17.58 & 11.72&19.33 &19.76 &14.68 &21.27 & 78 ms & 50 M\\
        IP-Net \cite{Wang2020IPNet} & \ding{55} & \ding{55} &19.56 &12.79 &21.58 &22.05 &15.77 &23.92 & - & - \\
        PPDM-Hourglass \cite{liao2019ppdm} & \ding{55} & \ding{55} & 21.73 & 13.78 & \textbf{24.10} & 24.58 & 16.65 & 26.84 & 71 ms & 195 M\\
        \midrule
        \textbf{DIRV (ours)} & \ding{55} & \ding{55} & \textbf{21.78} & \textbf{16.38} & 23.39 & \textbf{25.52} & \textbf{20.84} & \textbf{26.92} & \textbf{68 ms} & \textbf{13 M} \\
        \bottomrule
    \end{tabular}
    }
    \vspace{0in}
\end{table*}

\section{Experiments}
In this section, we carry out comprehensive  experiments to demonstrate the superiority of our proposed \textbf{DIRV}. Firstly, we introduce two benchmarks in Sec.~\ref{sec:dataset} and model implementation details in Sec.~\ref{sec:implement}. Then, we compare the performance of our model with other state-of-the-art approaches in Sec.~\ref{sec:result}.  Finally, effect of some crucial configurations are examined with ablation study in Sec.~\ref{sec:ablation}

\subsection{Dataset and Metric}
\label{sec:dataset}
\paragraph{Dataset}
We evaluate our method on two popular datasets: \textbf{V-COCO} \cite{gupta2015visual} and \textbf{HICO-DET} \cite{chao:iccv2015}. V-COCO dataset is a subset of COCO \cite{lin2014microsoft} with extra interaction labels. It contains 10,346 images (2,533 for training, 2867 for validation and 4,946 for testing). Each person in these images is annotated with 29 action categories, 4 of which (\textit{stand, smile, walk, run}) have no object. HICO-DET is a large dataset for HOI detection by augmenting HICO dataset \cite{chao:iccv2015} with instance bounding box annotations. This dataset includes 38,118 images for training and 9,658 images for testing. It is labelled with 600 HOI types over 117 verbs and 80 object categories.

\paragraph{Metric}
We adopt the popular evaluation metric for HOI detection: \textit{mean average precision (mAP)}. A prediction is true positive only when the HOI classification result is accurate as well as bounding boxes of human and object both have IoUs larger than 0.5 with reference to ground-truth. Specifically, we follow prior works to report \textit{Scenario 1 role mAP} on V-COCO dataset.

\subsection{Implementation Details}
\label{sec:implement}
For HOI detection, we use EfficientDet-d3 \cite{2019EfficientDet} as the backbone due to its effectiveness and efficiency. The backbone is pre-trained on COCO dataset. The \textit{instance classification and regression branches} are also initialized with the COCO pre-trained weight, which is frozen during training. We apply random flip and random crop data augmentation approaches to our model. Adam optimizer \cite{adam2014} is employed to optimize the loss function. We set the learning rate as 1e-4 with a batch size of 32. All experiments are carried out on NVIDIA RTX2080Ti GPUs.

\subsection{Results and Comparison}
\label{sec:result}
We compare our proposed \textbf{DIRV} with other state-of-the-art methods on V-COCO (Tab.~\ref{tab:vcoco}) and HICO-DET (Tab.~\ref{tab:hico}) datasets. It is noticeable that many state-of-the-art models utilize other additional features like human poses and language priors. These methods require additional data, annotations or models, which are quite exhaustive to collect. For fairness, we \textit{do not} take them (\textcolor{gray}{gray} ones in both Tab.~\ref{tab:vcoco},\ref{tab:hico}) into account in our comparison. What's more, unlike many existing two-stage approaches, our method does not rely on object proposals, which significantly elevates its compatibility.

For V-COCO dataset (Tab.~\ref{tab:vcoco}), we follow prior works to ignore the class \textit{point} since it has too few samples. Compared to prior arts, our approach outperforms them in accuracy significantly. It also has a fastest inference speed and a least parameter number.

For HICO-DET dataset (Tab.~\ref{tab:hico}), we report the results on
two different settings: Default and Known Objects. The interaction classification branch only classifies verb categories e.g. eating, which are associated with object categories e.g. apple based on the results of instance classification branch, as in \cite{2020UnionDet}. This classification strategy brings a more promising performance than directly recognizing the verb-object pair. The reason may be that it reduces the number of categories in interaction classification branch, which elevates the accuracy. What's more, it also saves the space overhead, allowing a larger batch size during training and improving the training stability. The results also demonstrate that our approach has a superiority in time and space complexity.

Two prior arts share some common insights with us. InteractNet \cite{Gkioxari_2018_CVPR} localizes objects based on single human appearance. UnionDet \cite{2020UnionDet} is another anchor-based one-stage HOI detection approach, focusing on \textit{union regions}. However, we surpass their performance by a large margin on both datasets, which proves the effectiveness of our concentration on interaction regions and our dense interaction region voting strategy.

In the supplementary materials, we show some qualitative results of our network, which is further analyzed with visualization.

\subsection{Ablation Study}
\label{sec:ablation}
In this section, we dig into the influence of different modules in our \textbf{DIRV}. For simplicity, all results here are for V-COCO dataset. Analysis of more components are available in the supplementary materials.

\begin{table}[ht!]
    \centering
    \caption{\textbf{Interaction Region Overlapping Thresholds:} $t_u,t_h,t_o$ denote the thresholds in Eq.~\ref{eq:overlapping}. The interaction regions become denser as these three thresholds decrease.}
    \begin{tabular}{cccc}
        \toprule
         $\mathbf{t_h}$ & $\mathbf{t_o}$ & $\mathbf{t_u}$ & $\mathbf{mAP}_{role}$  \\
         \midrule
         $0.5$ & $0.5$ & $0.5$ & $55.0$\\
         $0.25$ & $0.25$ & $0.5$ & $55.2$ \\
         $0.25$ & $0.25$ & $0.25$ & $\textbf{56.1}$ \\
         \bottomrule
    \end{tabular}
    \label{tab:overlap}
\vspace{0in}
\end{table}

\paragraph{Interaction Regions Overlapping Thresholds}
We set interaction regions in a dense manner for human-object pairs. The overlapping thresholds in Eq.~\ref{eq:overlapping} is examined in this part. Results in Tab.~\ref{tab:overlap} certificate this dense manner, which can make full use of the visual features.

\begin{figure}[t]
    \centering
    \includegraphics[width=0.8\linewidth]{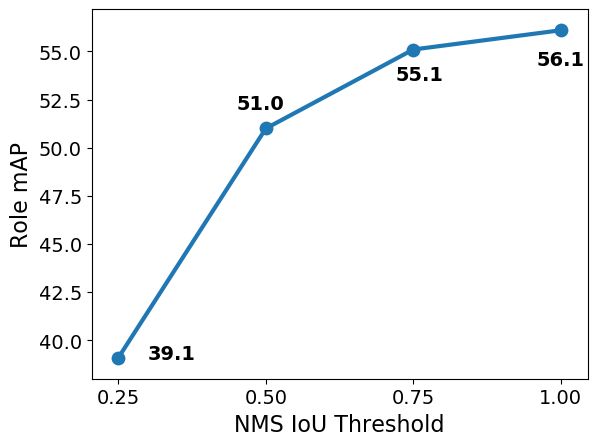}
    \caption{\textbf{Ablation Study for Voting Strategy:} The $mAP_{role}$ increases as the IoU threshold for NMS grows. There is actually no NMS when IoU threshold is $1$.}
    \label{fig:nms_chart}
\end{figure}

\paragraph{Voting Strategy}
We examine the superiority of our voting strategy by adding a NMS module for interaction regions, which weakens the effect of voting. In Fig.~\ref{fig:nms_chart}, we set different IoU thresholds for NMS and the performance drops as the value of those thresholds decreases (when IoU threshold is $1$, NMS takes no effect).  It reveals that interaction regions of different scales all contribute to the final detection though some of their classification scores may not be very high.

\begin{table}[ht!]
    \centering
     \caption{\textbf{Loss Function for Interaction Classification}}
    \begin{tabular}{lc}
        \toprule
        \textbf{Loss Function} & $\mathbf{mAP}_{role}$ \\
        \midrule
         Focal Loss \cite{DBLP:journals/corr/abs-1708-02002}  & $54.8$  \\
         Foreground Loss \cite{2020UnionDet} & $54.0$ \\
         Ignore Loss (ours) & $\textbf{56.1}$\\
         \bottomrule
    \end{tabular}

    \label{tab:loss}
\end{table}

\paragraph{Ignorance Loss}
We look into the effect of loss function in \textit{interaction classification branch}. We test the performance with vanilla focal loss, foreground loss in \cite{2020UnionDet} and our proposed \textit{ignorance loss}. Results in Tab.~\ref{tab:loss} verify our superiority since it can help dealing with region overlapping and missed positive labels.

\paragraph{Backbone}

We apply a novel backbone   \cite{2019EfficientDet} to our model, which has never been utilized for HOI detection.

We separately carry out experiments with EfficientDet-d1, d2, d3 and d4. To our surprise, we find that the heavier backbone doesn't certainly lead to better HOI detection performance, according to the results in Tab.~\ref{tab:backbone}.

We also reproduce another anchor-based one-stage algorithm UnionDet \cite{2020UnionDet} with EfficientDet-d3 backbone. Results in Tab.~\ref{tab:backbone} reveals that our DIRV surpasses it because of our novel design in methodology, instead of the backbone improvement.

\begin{table}[htb!]
    \centering
    \caption{\textbf{Ablation Study for Backbones:} We compare the performance of our DIRV and another anchor-based method UnionDet \cite{2020UnionDet} with different backbones.}
    \label{tab:backbone}
    \begin{tabular}{lccc}
        \toprule
        \textbf{Method} & \textbf{Backbone (Params)} & $\textbf{mAP}_{role}$\\
        \midrule
        \multirow{2}{*}{UnionDet} & ResNe50-FPN (34M) & 47.5\\
         & EfficientDet-d3 (12M) & 49.2 \\
        \midrule
        \multirow{4}{*}{DIRV (ours)} & EfficientDet-d1 (6.6M) & 46.8 \\
        & EfficientDet-d2 (8.1M) & 51.3\\
        & EfficientDet-d3 (12M) & \textbf{56.1}\\
        & EfficientDet-d4 (21M) & 54.3\\
        \bottomrule
    \end{tabular}
\end{table}

% move to supplementary materials
% \subsubsection{Network Backbone}
% Since our HOI detection depends on the visual features extracted by backbone network, we want to understand how the backbone affects final detection. In Fig.~\ref{fig:backbone}, we apply EfficientDet \cite{2019EfficientDet} with different scales. To our surprise, there exists no positive correlation between HOI detection performance and backbone performance on object detection. We attribute it to that advanced object detection networks may extract high-level visual features specific to object detection in FPN, where some information of visual relationship is lost.

\section{Conclusion}
In this paper, we present a novel one-stage HOI detection framework. It detects HOI in an intuitive manner by concentrating on the \textit{interaction regions}. To compensate for the detection flaws of single interaction region, a \textit{voting strategy} is applied as an alternative to conventional NMS. Our method outperforms all existing approaches without any additional features or proposals. Due to the one-stage structure and simple network architecture, our method reaches a very high efficiency with least model parameters compared to other state-of-the-art approaches. In the future, we will try to incoporate the part-level knowledge~\cite{li2019hake} into our framework.

\paragraph{Acknowledgement} This work is supported in part by the National Key RD Program of China, No. 2017YFA0700800, National Natural Science Foundation of China under Grants 61772332, Shanghai Qi Zhi Institute, SHEITC(2018-RGZN-02046) and Baidu Fellowship.

{\small
\bibliographystyle{ieee}
\bibliography{egbib}
}

\newpage

\begin{appendix}

In this supplement, we provide more analysis and experiments not included in the main paper due to space limitation. They are listed as follows:
\begin{itemize}
\item Analysis of performance and efficiency is given in Sec.~\ref{sec:performance}. We compare our method with other existing ones.
\item We show some qualitative results of our proposed \textit{interaction regions} in Sec.~\ref{sec:qualitive}
\item More ablation studies are conducted to examine some components of our DIRV in Sec.~\ref{sec:ablation_s}.
\item We visualize some examples of HOI detection in various cases to analyze the generality of our DIRV in Sec.~\ref{sec:visual}.
\end{itemize}

\section{Performance and Efficiency}
\label{sec:performance}
As mentioned in the main paper, our DIRV surpasses other state-of-the-art approaches in accuracy with both fewer parameters and faster inference speed.

For parameter counting, we follow the estimation strategy in \cite{DBLP:journals/corr/abs-1904-03181} to calculate the parameter number of iCAN \cite{gao2018ican} and TIN \cite{li2019transferable}. Similar estimation is also applied to UnionDet \cite{2020UnionDet}, InteractNet \cite{Gkioxari_2018_CVPR} and IP-Net \cite{Wang2020IPNet} since the authors did not provide open-source codes. For VSGNet \cite{2020VSGNet} and PMFNet \cite{wan2019pose}, parameters are counted based on the open-source codes.

For time estimation, we consider the sum of the object detection time and HOI detection time for those two-stage approaches, including iCAN \cite{gao2018ican}, TIN \cite{li2019transferable}, InteractNet \cite{Gkioxari_2018_CVPR} and VSGNet \cite{2020VSGNet}. We run different models on a NVIDIA RTX2080Ti GPU and some results are referred from other published work \cite{2020UnionDet,liao2019ppdm}.

\begin{figure}[t!]
    \centering
    \subfigure[\textbf{accuracy} \textit{vs} \textbf{speed}]{
        \centering
        \includegraphics[width=0.46\linewidth]{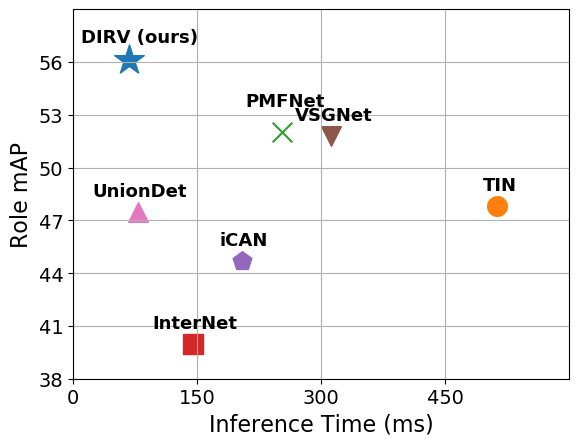}
    }
    \subfigure[\textbf{accuracy} \textit{vs} \textbf{size}]{
        \centering
        \includegraphics[width=0.46\linewidth]{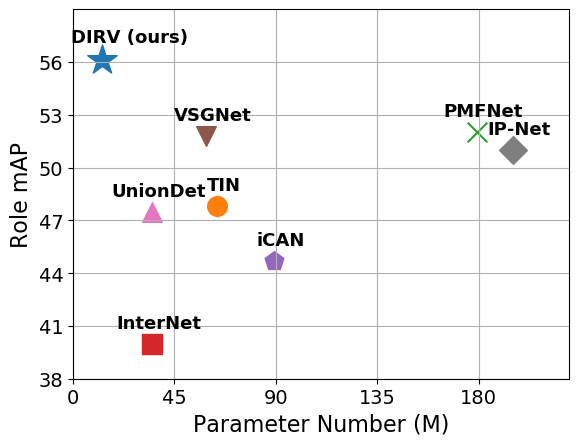}
        \label{fig:interaction region overlap}
    }
    \caption{\textbf{mAP versus Inference Time/Parameter Number on V-COCO dataset:} Our proposed DIRV reaches a new state-of-the-art 56.1 $mAP_{role}$ with fastest inference time (68 ms) and fewest parameters (13M) compared with previous methods.}
    \label{fig:mAP_time_params}
\end{figure}

In Fig.~\ref{fig:mAP_time_params}, we illustrate the performance of different models versus inference time and parameter number separately on V-COCO dataset. It is apparent that our DIRV outperforms others remarkably with a significant superiority in both time and space efficiency.

\begin{figure}[htb!]
    \centering
    \includegraphics[width=0.31\linewidth]{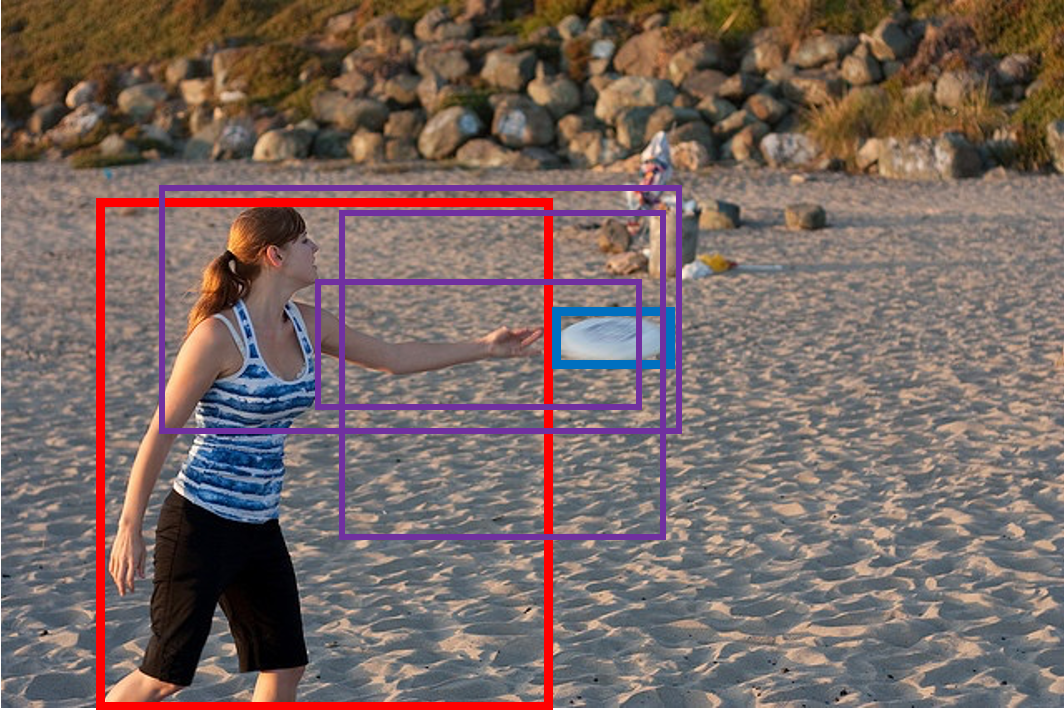}
    \includegraphics[width=0.31\linewidth]{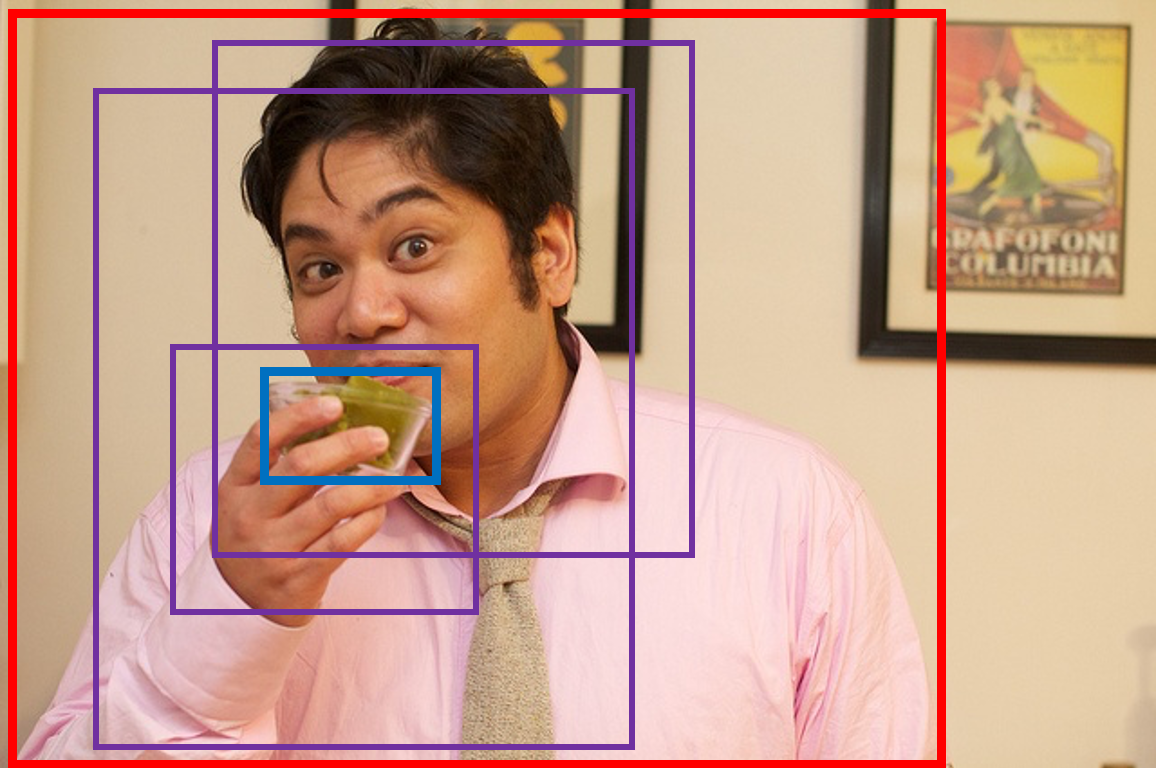}
    \includegraphics[width=0.31\linewidth]{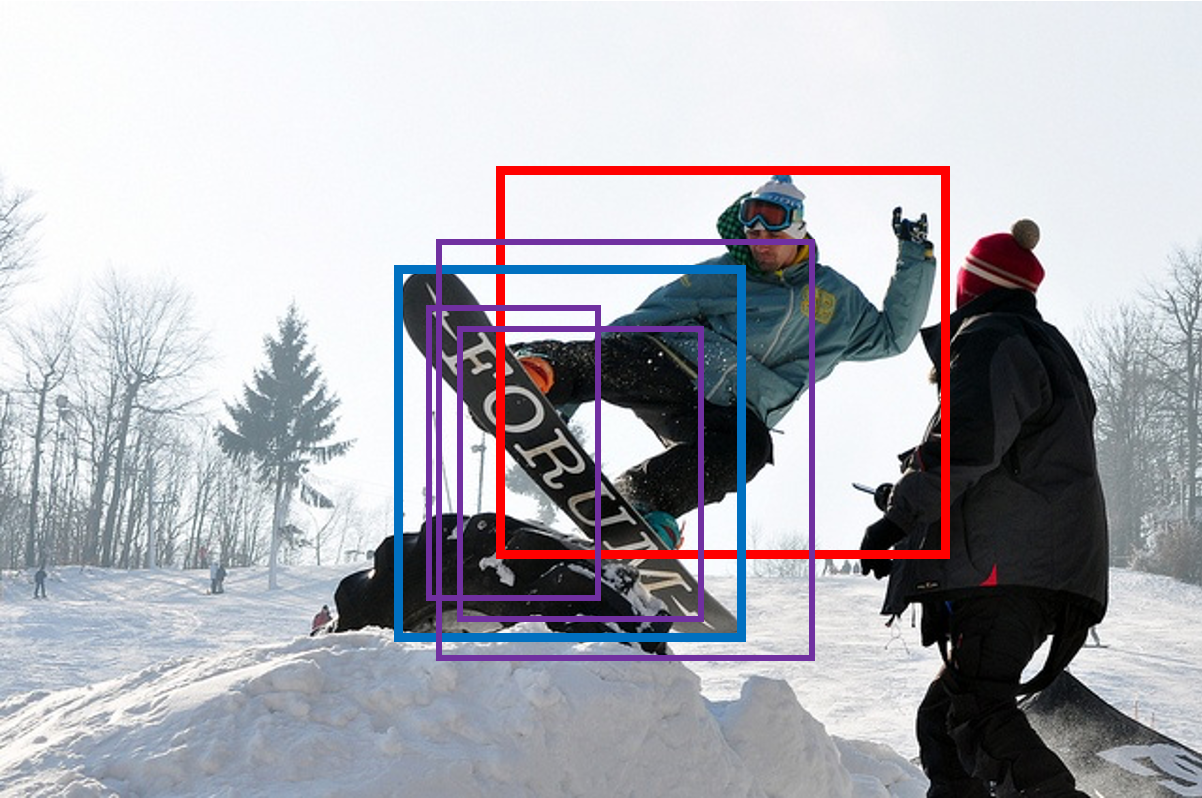}

    \includegraphics[width=0.31\linewidth]{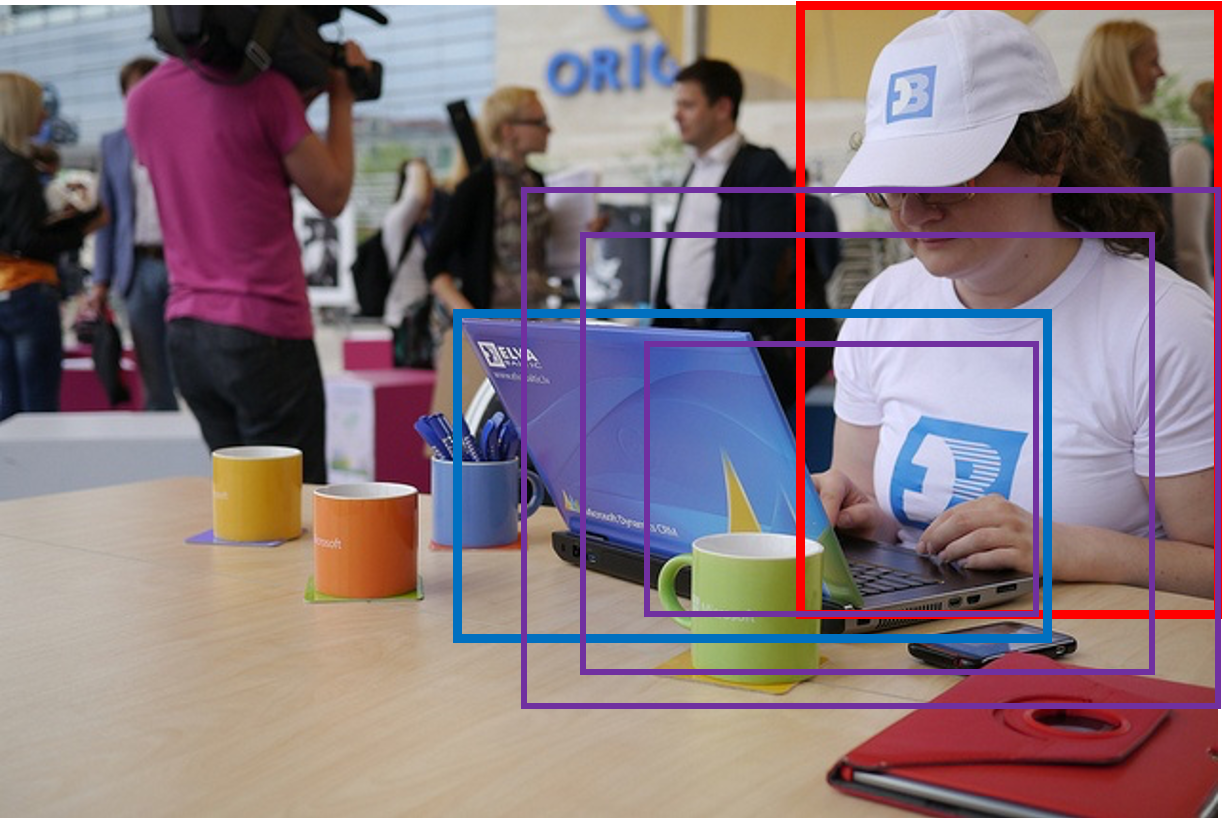}
    \includegraphics[width=0.31\linewidth]{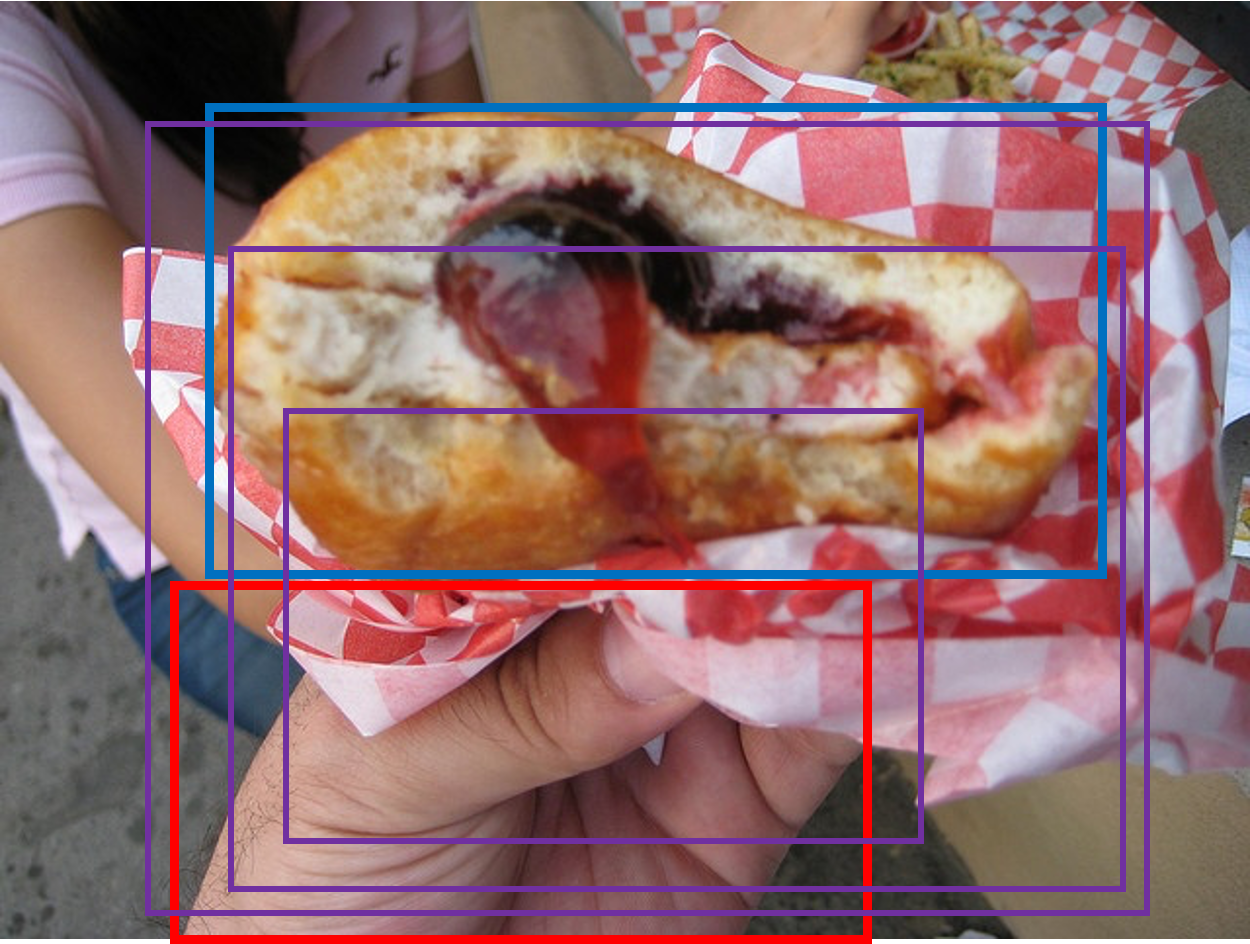}
    \includegraphics[width=0.31\linewidth]{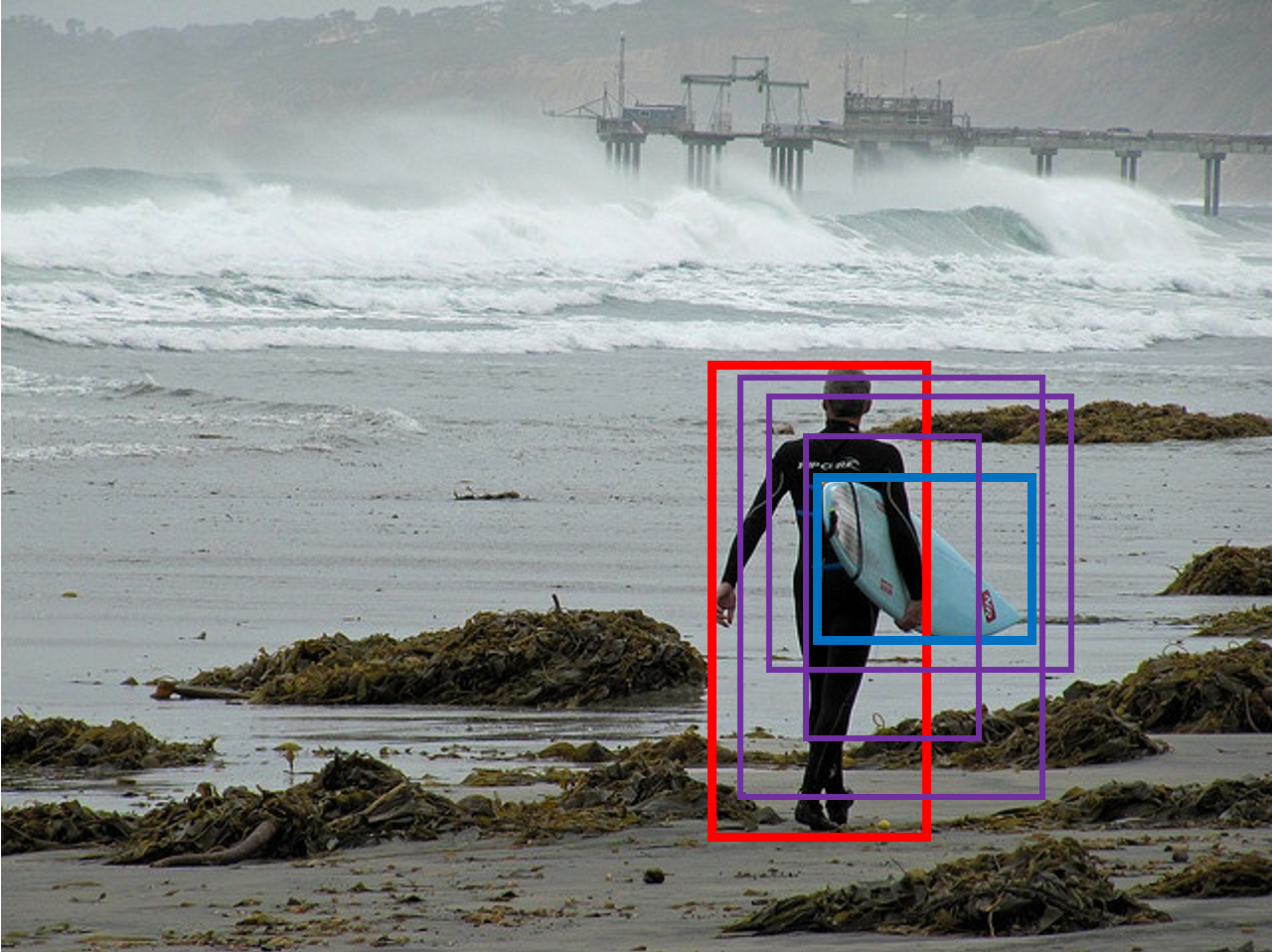}

    \subfigure{
        \includegraphics[width=0.23\linewidth]{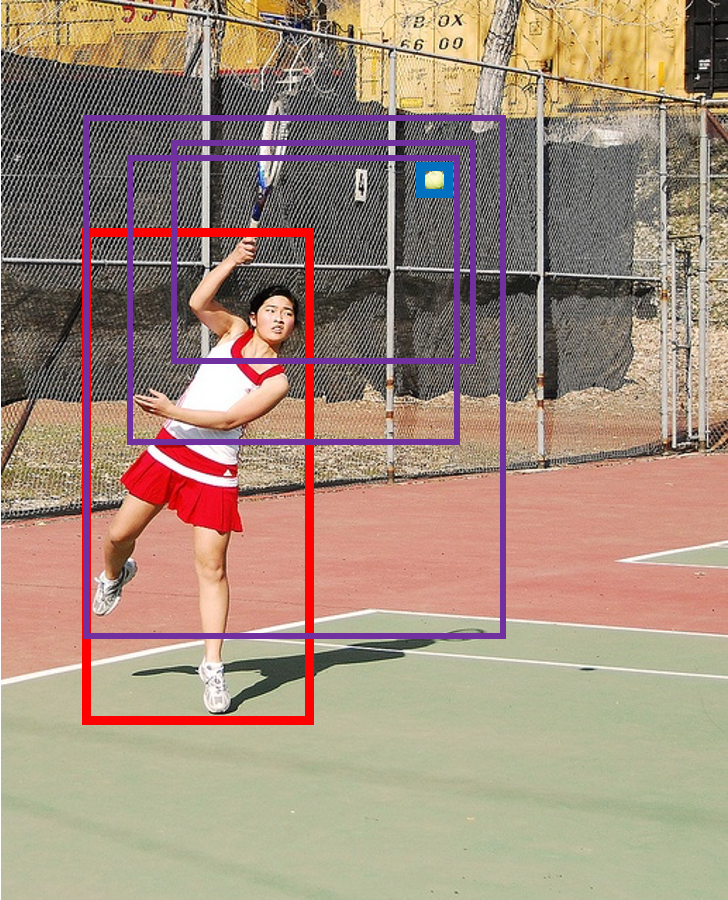}
        \includegraphics[width=0.23\linewidth]{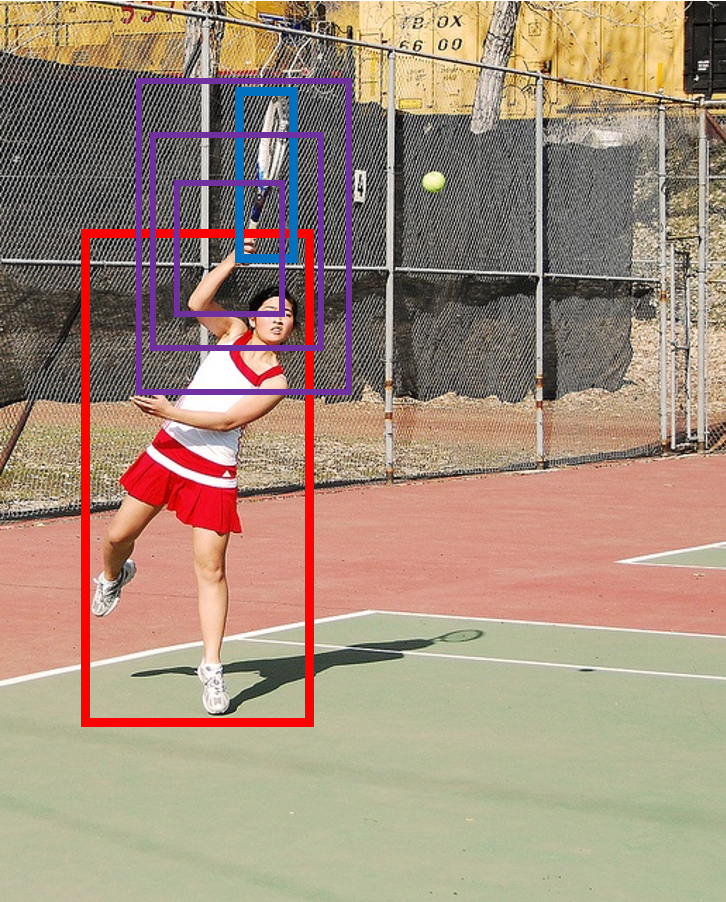}
    }
    \subfigure{
        \includegraphics[width=0.21\linewidth]{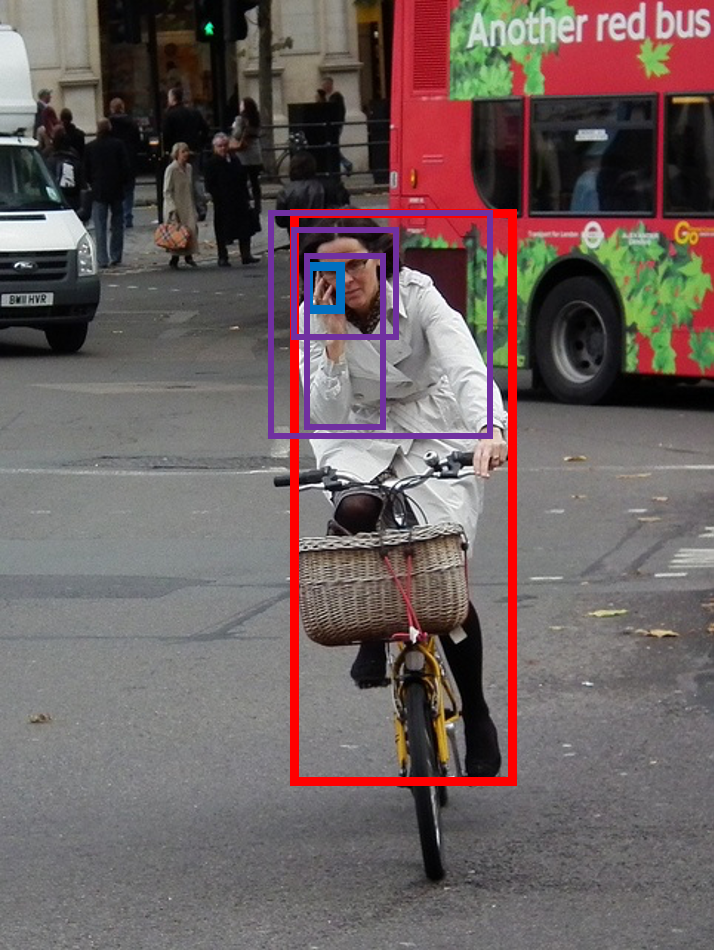}
        \includegraphics[width=0.21\linewidth]{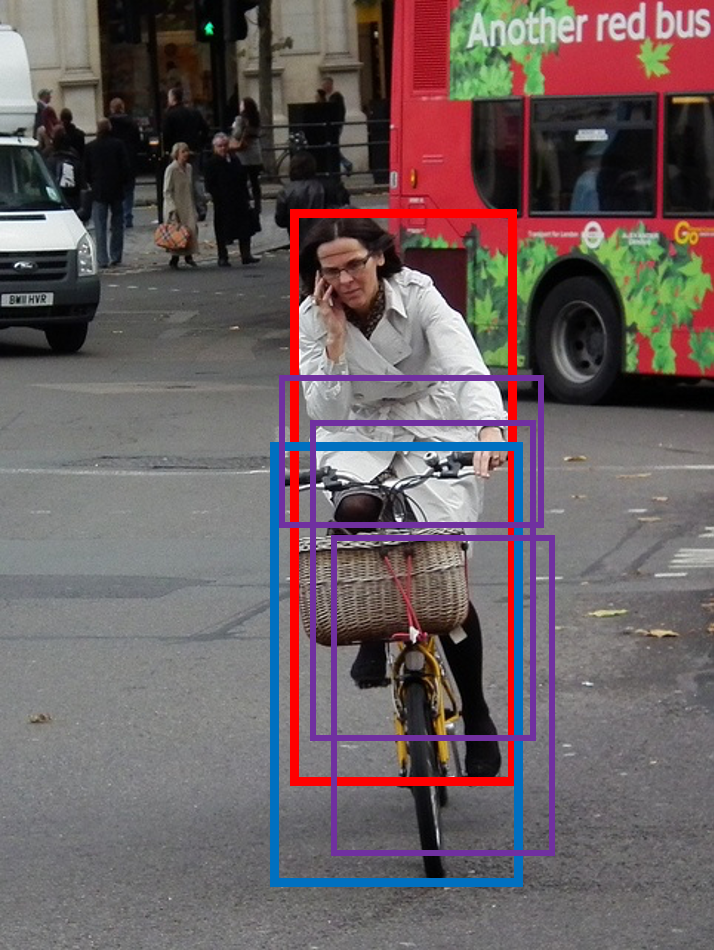}
    }
    \caption{\textbf{Examples of Interaction Regions:} \textit{Red} and \textit{blue} rectangles respectively denote the interacting humans and objects. Interaction regions are drawn in \textit{purple}. During training, interaction regions are actually much denser than these illustrations.}
    \label{fig:regions}
\end{figure}

\section{Analysis of Interaction Regions}
\label{sec:qualitive}

We provide more examples of interaction regions with different scales in Fig.~\ref{fig:regions}, where each interaction region is associated with a specific human-object interaction. These interaction regions are composed of parts of the human, object and context, containing visual features essential for HOI detection.

\begin{figure}[htb!]
    \centering
    \includegraphics[width=0.7\linewidth]{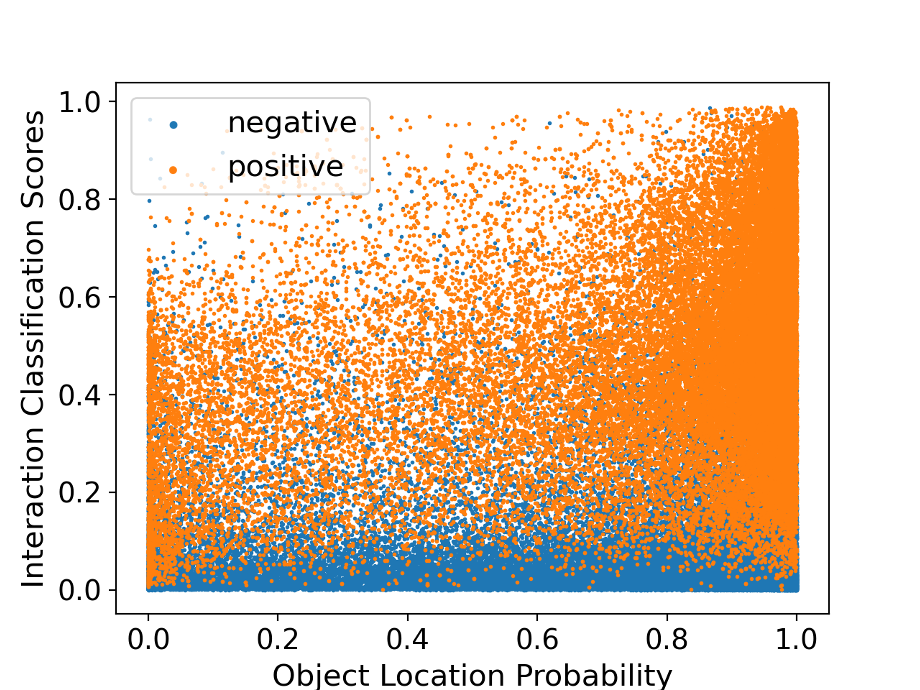}
    \caption{\textbf{Scores of Interaction Regions:} It shows \textit{object location probability} $p_{a_j}(x_o^{a_j},y_o^{a_j})$ and \textit{interaction classification scores} $s_{a_j}^{inter}$ of interaction regions for different human-object pairs. Interaction regions of positive and negative pairs are marked in \textit{orange and blue} separately.}
    \label{fig:region_score}
\end{figure}

Fig.~\ref{fig:region_score} visualizes the \textit{interaction classification scores} $s_{a_j}^{inter}$ versus \textit{object location probability} $p_{a_j}(x_o^{a_j},y_o^{a_j})$ of different interaction regions for some human-object pairs. We mark positive (with interactions) and negative (w/o interactions) pairs with different colors. There are two hints in this image. Firstly, positive pairs are predicted with notably higher interaction classification scores in most interaction regions since our interaction regions capture the most crucial visual features for interactions. Secondly, object location probability for positive regions is not certainly very high. The relative spatial relationship is reflected from some very subtle visual features, which are hard to be completely discovered from a single interaction region. This corroborates the necessity of our voting strategy.

\begin{figure}[htb!]
    \centering
    \subfigure{
        \includegraphics[width=0.23\linewidth]{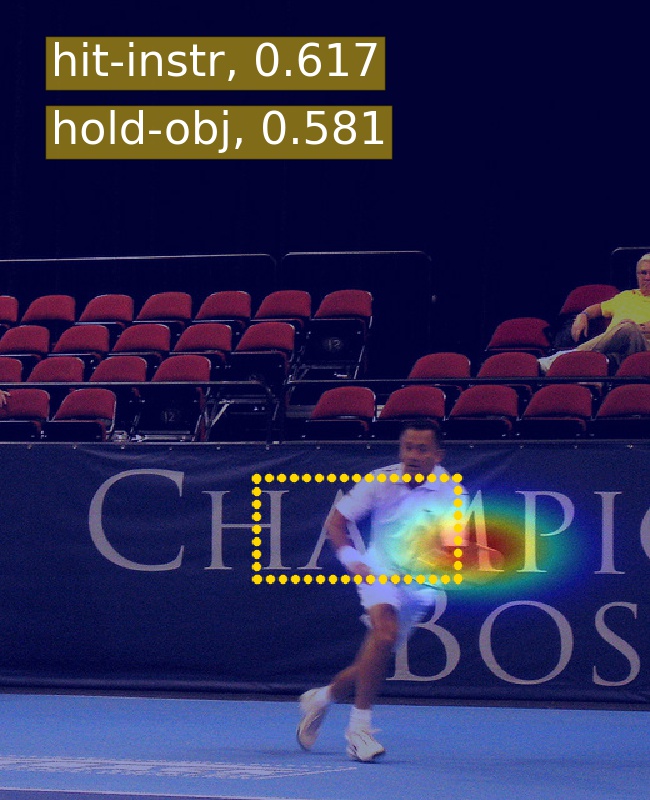}
        \includegraphics[width=0.23\linewidth]{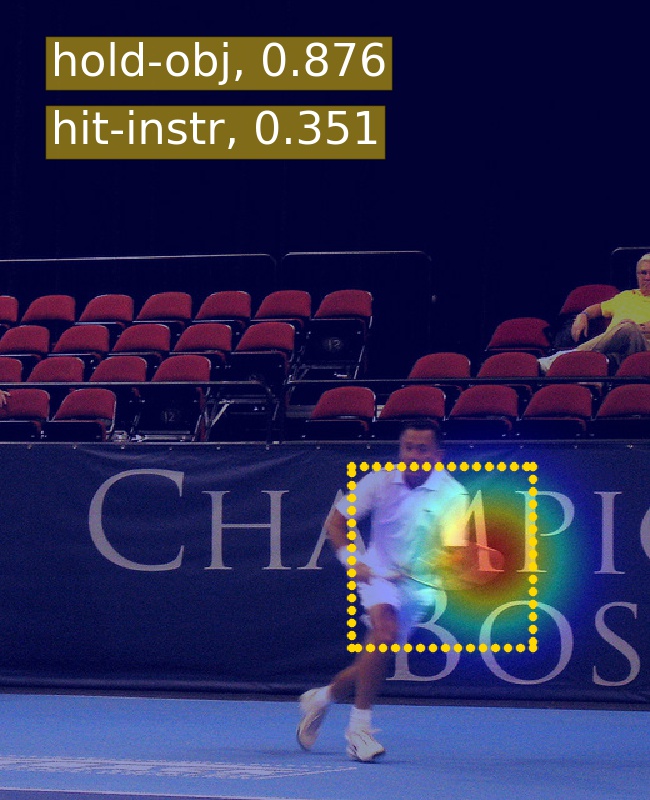}
        \includegraphics[width=0.23\linewidth]{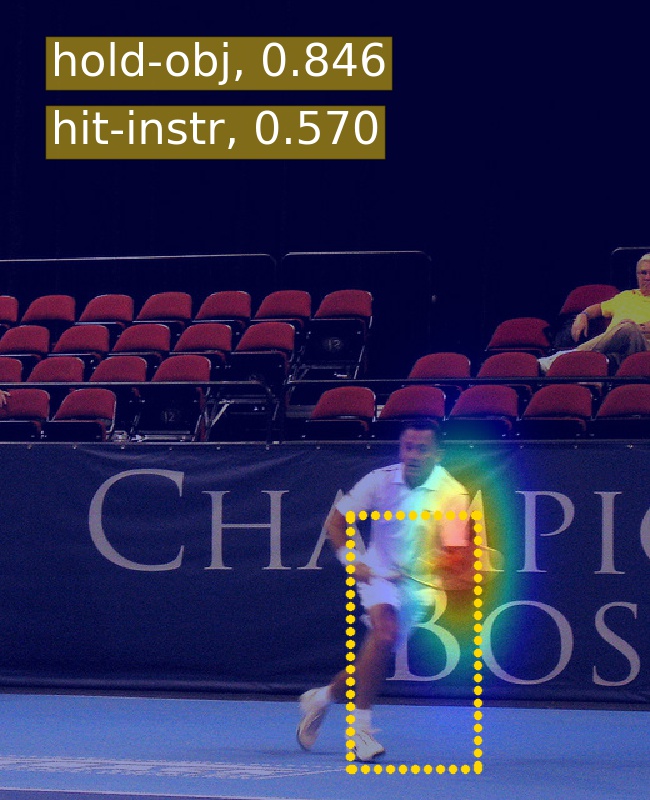}
        \includegraphics[width=0.23\linewidth]{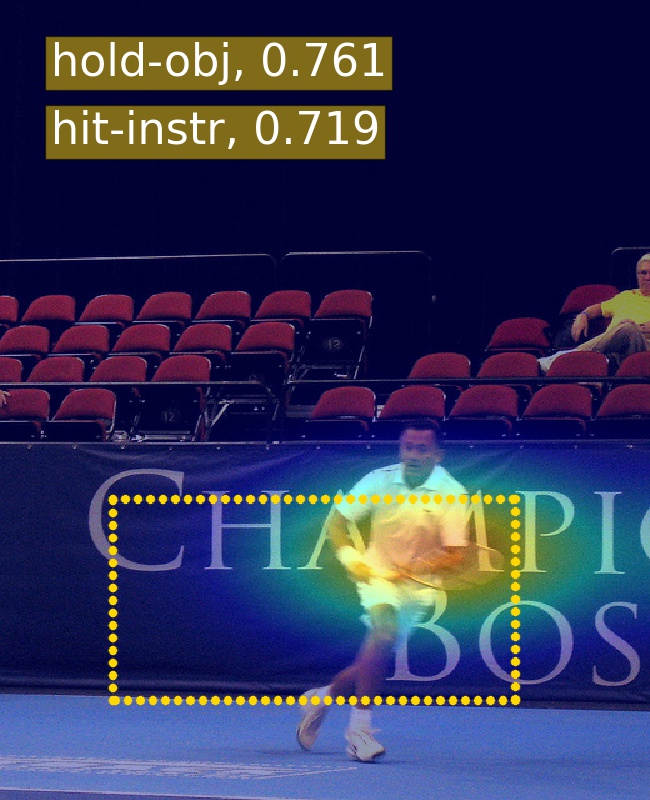}
    }
    \subfigure{
        \includegraphics[width=0.31\linewidth]{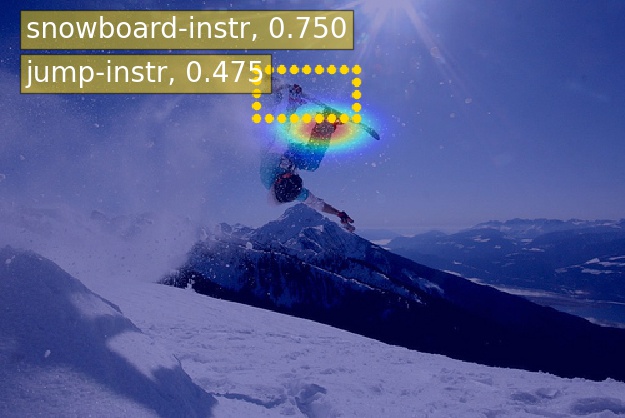}
        \includegraphics[width=0.31\linewidth]{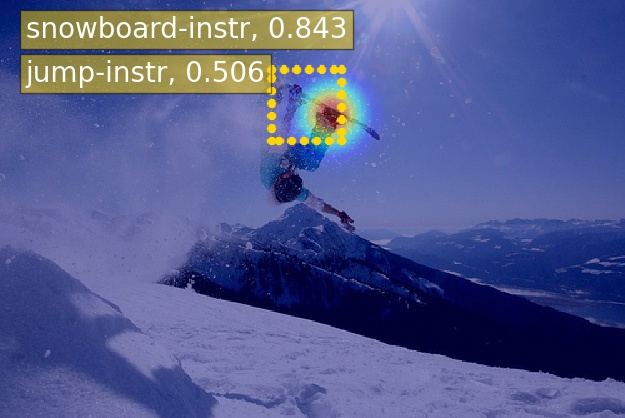}
        \includegraphics[width=0.31\linewidth]{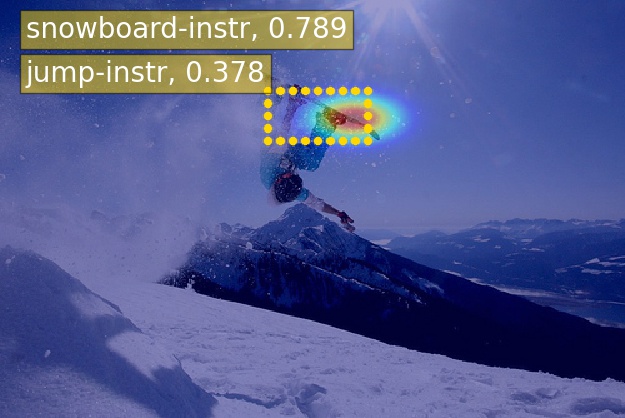}
    }
    \caption{\textbf{Detection Result of Single Interaction Region:} \textit{Yellow dotted} rectangles denote the interaction regions. We visualize the object location distribution and list the classification results for each interaction region. Note that here \textit{hit-instr} means \textit{hit with instrument} in the upper line, whose target object is the racket.}
    \label{fig:single_res}
\end{figure}

As a supplement, we also illustrate \textit{interaction detector} results of some single interaction regions in Fig.~\ref{fig:single_res}. Most single regions can fetch satisfying classification results but there exist clear errors in object localization. However, since the errors are distributed in all directions uniformly, they are counterbalanced through voting.

\section{More Ablation Studies}
\label{sec:ablation_s}

We add three extra ablation studies on V-COCO dataset here. Firstly, we verify the significance of our \textit{interaction detector}, which serves as the key of our DIRV. Then, we examine the effect of scores from \textit{instance detector}. Finally, we consider different values for the standard deviation $\sigma$ of the 2-d Gaussian distribution in Eq.~\ref{eq:gaussian} of the main paper.

\subsection{Interaction Detector}
Since the results can be derived from the \textit{instance detector} alone, we try to eliminate the whole \textit{interaction detector}, i.e. set $s_{h,o}^{fuse}=1$. In this case, we have
\begin{equation}
    S_{h,o} = s_h\cdot s_o\cdot (s_h^{act}+s_o^{act})
\end{equation}
in place of Eq.~\ref{eq:score} in the main paper. Results in Tab.~\ref{tab:instance} witness a dramatic drop of 15.5 mAP, which verifies the indispensability of our novel \textit{interaction detector}.

\begin{table}[htb!]
    \centering
    \caption{\textbf{Effect of Each Score for  HOI Prediction:} We examine the role of each score ($s_h, s_o, s_h^{act}, s_o^{act}, s_{h,o}^{fuse}$) in Eq.~\ref{eq:score}. The lack of \textit{interaction detector} brings significant performance drop, while removing instance classification scores or instance action classification scores only leads to limited performance drop.
    }
    \label{tab:instance}
    \begin{tabular}{lcccc}
        \toprule
        \multirow{2}{*}{\textbf{Method}} & \multicolumn{3}{c}{\textbf{Scores}} & \multirow{2}{*}{\textbf{Method}}\\
         & $s_h,s_o$ & $s_h^{act},s_o^{act}$ & $s_{h,o}^{fuse}$ & \\
        \midrule
        \multirow{4}{*}{DIRV} & \ding{51} & \ding{51} & & 40.6\\
        &\ding{51} &  &\ding{51} & 53.8\\
        & & \ding{51} &  \ding{51}& 53.6\\
        & \ding{51} & \ding{51} &\ding{51} & \textbf{56.1}\\
        \bottomrule
    \end{tabular}
\end{table}

\subsection{Instance Scores}
In Eq.~\ref{eq:score} of the main paper, the instance classification scores $s_h,s_o$ and instance action classification scores $s_h^{act},s_o^{act}$ from the \textit{instance detector} have an auxiliary influence on the final HOI score of a human-object pair. We analyze their exact roles by separately setting them as a fixed value $1$ when making the prediction. The results are shown in Tab.~\ref{tab:instance}. Although these scores are beneficial for the prediction, their effect is limited especially compared to the scores $s_{h,o}^{fuse}$ from the \textit{interaction detector}. This result reversely verifies the effectiveness of our \textit{interaction detector}.

% \begin{table}[htb!]
%     \centering
%     \caption{\textbf{Results with Instance Detector Alone:} The lack of interaction detector brings significant performance drop.
%     % And EfficientDet backbone only leads to limited improvement compared to ResNet-50.
%     }
%     \label{tab:instance}
%     \begin{tabular}{lccc}
%         \toprule
%         \textbf{Method} & \textbf{Backbone} & $\textbf{mAP}_{role}$\\
%         \midrule
%         % Instance Detector & ResNet-50-FPN & 38.4 \\
%         Instance Detector & EfficientDet-d3 & 40.6\\
%         % \midrule
%         DIRV & EfficientDet-d3 & \textbf{56.1}\\
%         \bottomrule
%     \end{tabular}
% \end{table}

% \subsection{Backbone}

% In our main paper, ablation study in Sec. 4.4 has examined the significance of our dense interaction regions, voting strategy and ignorance loss separately in three sub-sections. Further, we want to ensure that our improvement doesn't come from the backbone solely. In Tab.~\ref{tab:instance}, we compare the performance of two baselines with only the instance detector. They are equipped with two different backbones: ResNet-50-FPN and EfficientDet-d3. The former result comes from this paper \cite{2020UnionDet}. It is noticeable that the EfficientDet backbone only brings a modest elevation compared to the common used ResNet-50-FPN.

\subsection{Standard Deviation for Location Distribution}
\begin{figure}[htb!]
    \centering
    \includegraphics[width=0.8\linewidth]{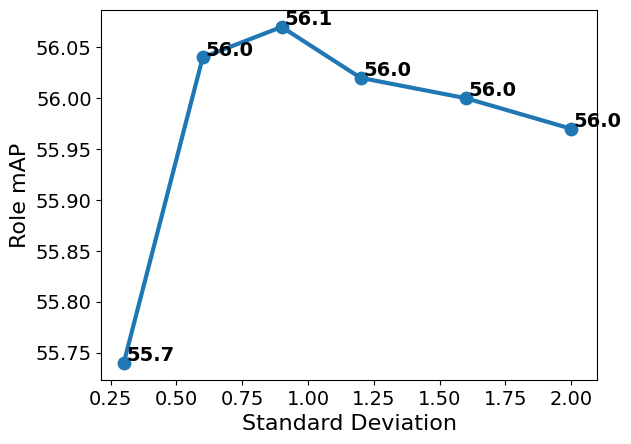}
    \caption{\textbf{Ablation Study for Standard Deviation $\sigma$ of Object Location Distribution}}
    \label{fig:sigma}
\end{figure}

In this part, we analyze the hyper-parameter $\sigma$ for the Gaussian distribution of the relative object location (Eq.~8 in the main paper). We find that the model performance is not very sensitive to this standard deviation, as is shown in Fig.~\ref{fig:sigma}. It shows the reliability and robustness of our interaction region prediction and voting strategy.

\begin{figure*}
    \centering
    \subfigure[large humans or objects]{
        \centering
        \includegraphics[width=0.9\linewidth]{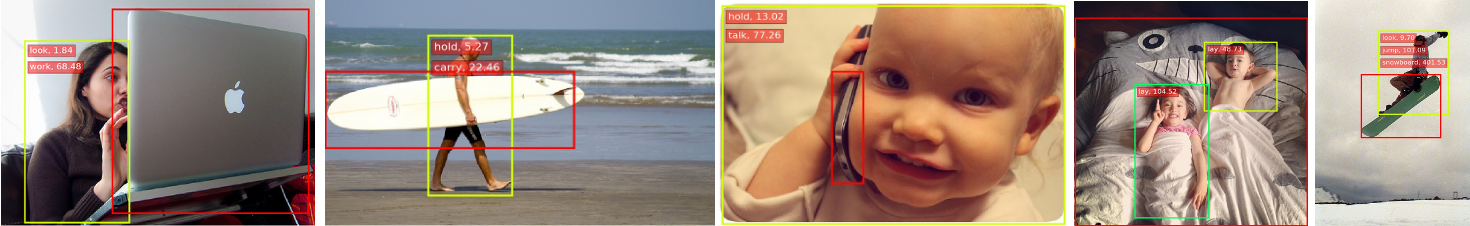}
    }
    \subfigure[small humans or objects]{
        \centering
        \includegraphics[width=0.9\linewidth]{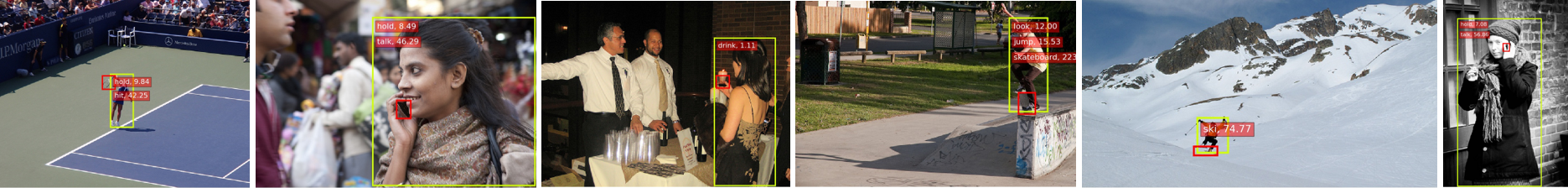}
    }
    \subfigure[humans remote from target objects]{
        \centering
        \includegraphics[width=0.9\linewidth]{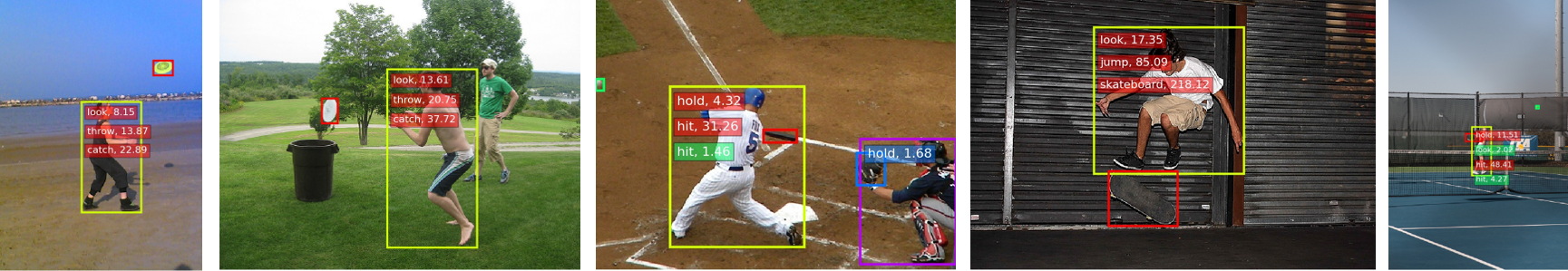}
    }
    \subfigure[humans close to target objects]{
        \centering
        \includegraphics[width=0.9\linewidth]{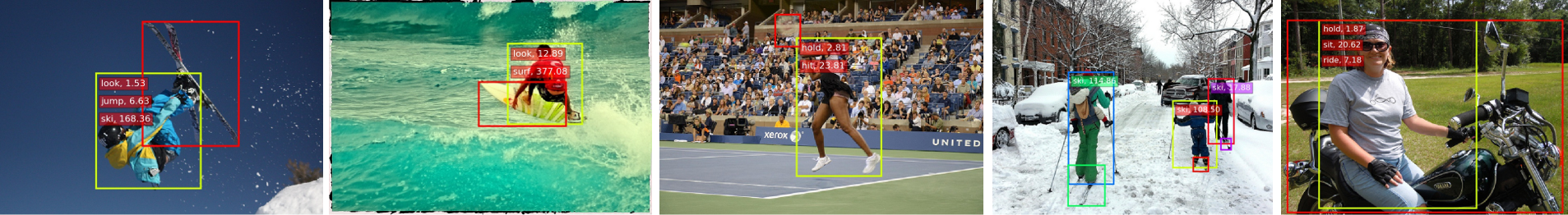}
    }
    \subfigure[humans interacting with multiple objects]{
        \centering
        \includegraphics[width=0.9\linewidth]{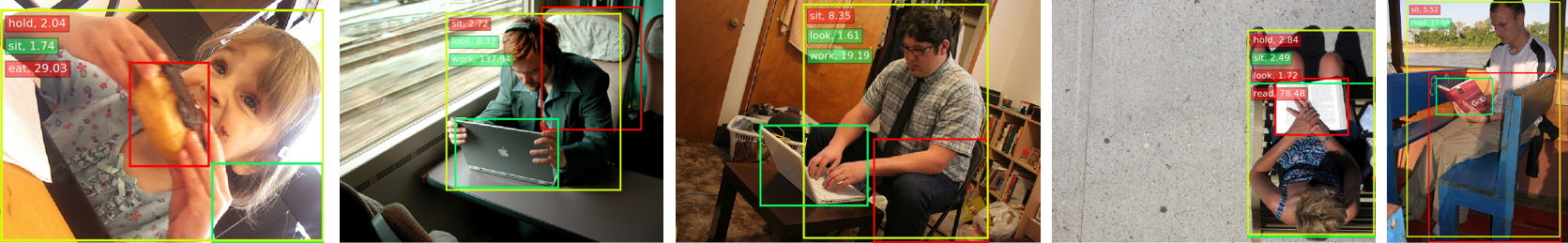}
    }
    \subfigure[different HOI pairs overlapping with each other]{
        \centering
        \includegraphics[width=0.9\linewidth]{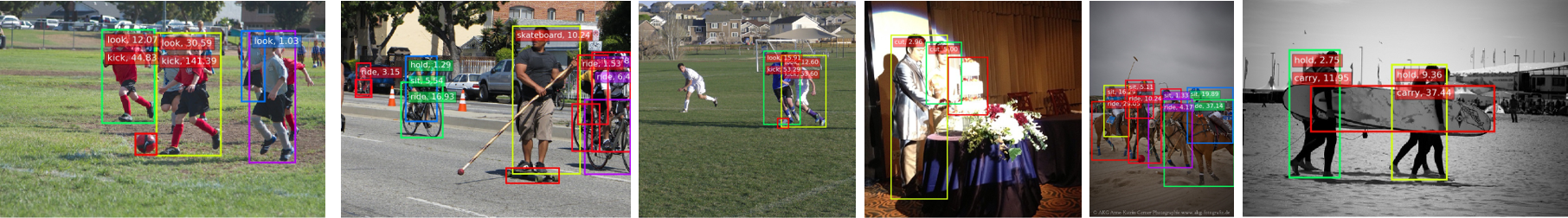}
    }
    \subfigure[incomplete humans or objects]{
        \centering
        \includegraphics[width=0.9\linewidth]{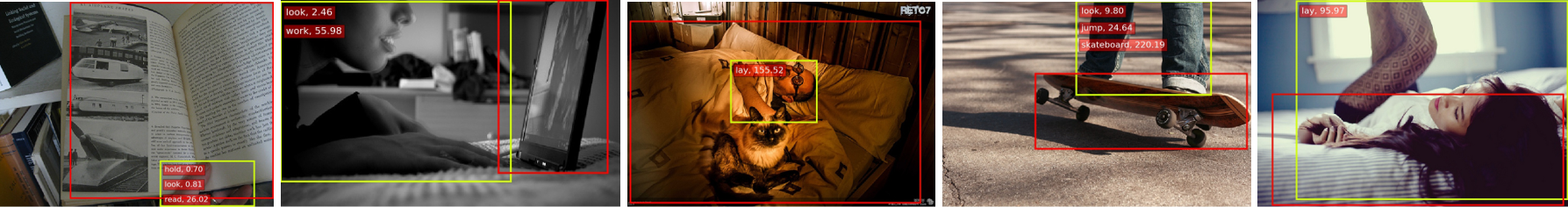}
    }
    \caption{\textbf{Visualization of Detection Results}}
    \label{fig:visual_detection}
\end{figure*}

\section{Detection Visualization}
\label{sec:visual}
We present some visualization of DIRV results in Fig.~\ref{fig:visual_detection}. For each human, the corresponding interaction labels and target objects are displayed in the same color. We mainly pay attention to examples with different characteristics. In these examples, our proposed DIRV deals with various situations very well, despite their special difficulties as follows.

For humans and objects vary in different sizes, our dense interaction regions can easily capture visual features of different scales, resulting in high confidence and accuracy.

For objects remote from humans, there exist less interactive clues in interaction regions, which makes the prediction harder than close human-object pairs. Despite the overall great performance, several ambiguous interactions (\textit{e.g.} catch and throw) share some common features, bringing possible detection flaws. Multiple interactions of same or different humans may share overlapping interaction regions, which generates potential confusion during training. Yet, our proposed DIRV solved these problems well, obtaining satisfying performance in these cases.

Since our interaction regions focus on parts of humans or objects most essential for interaction, incomplete human or object instances can hardly have any negative influence on the detection.

All the examples above verify the strong generality of our proposed approach. We are looking forward to its wide application in different practical applications.

\end{appendix}
\end{document}